\documentclass[10pt,twocolumn,letterpaper]{article}

\usepackage{cvpr}

\usepackage{times}
\usepackage{epsfig}
\usepackage{graphicx}
\usepackage{amsmath}
\usepackage{amssymb}
\usepackage{pifont}
\usepackage{booktabs}
\usepackage{multirow}
\usepackage{arydshln}
\usepackage{lipsum}  
\usepackage{enumitem}
\usepackage[table]{xcolor}
\usepackage{bbm}
\usepackage{arydshln}
\usepackage{pifont}
\usepackage{epsfig}
\usepackage{graphicx}
\usepackage{adjustbox}
\usepackage{array}
\usepackage{booktabs}
\usepackage{colortbl}
\usepackage{float,wrapfig}
\usepackage{hhline}
\usepackage{multirow}
\usepackage{amsmath,amsfonts,amsthm,amssymb}
\usepackage{bm}
\usepackage{nicefrac}
\usepackage{microtype}
\usepackage{enumerate}
\usepackage{changepage}
\usepackage{extramarks}
\usepackage{fancyhdr}
\usepackage{lastpage}
\usepackage{setspace}
\usepackage{soul}
\usepackage{xspace}
\usepackage{booktabs}
\usepackage{longtable}
\usepackage{arydshln}
\usepackage{amssymb}
\usepackage{pifont}
\usepackage{xcolor}
\usepackage[T1]{fontenc}
\usepackage[font=small,labelfont=bf,tableposition=top]{caption}
\DeclareCaptionLabelFormat{andtable}{#1~#2  \& \tablename~\thetable}
\usepackage{capt-of}
 \usepackage{relsize}
\usepackage{epigraph}
\usepackage{diagbox}
\usepackage{makecell}
\usepackage{xr}
\usepackage{xr-hyper}
\usepackage{lipsum}  

\definecolor{darkgreen}{rgb}{0.0, 0.4, 0.23}
\definecolor{darkred}{rgb}{0.6, 0.16, 0.25}
\definecolor{darkblue}{rgb}{0.0, 0.16, 0.75}
\definecolor{maroon}{rgb}{0.76, 0.13, 0.28}
\definecolor{codegreen}{rgb}{0,0.6,0}
\definecolor{codegray}{rgb}{.95,.95, .95}
\definecolor{codepurple}{rgb}{0.58,0,0.82}
\definecolor{backcolour}{rgb}{0.95,0.95,0.92}
\definecolor{p}{rgb}{148,2,209}
\definecolor{darkbrown}{rgb}{0.3, 0.15, 0}
\definecolor{ddarkgreen}{rgb}{0.0, 0.4, 0.23}

\newcommand{\vitcap}{{\normalsize ViT}{\normalsize CAP}\;}
\newcommand{\vitcapp}[0]{{\small ViT}{\small CAP}\;}
\newcommand{\coco}[0]{{{COCO-caption}}\;}

\usepackage{transparent}

\usepackage[pagebackref=true,breaklinks=true,colorlinks,bookmarks=false]{hyperref}
\hypersetup{
colorlinks=true,
linkcolor=darkred,
filecolor=magenta,      
urlcolor=darkred,
citecolor=darkgreen,
}   
\usepackage{enumitem}
\usepackage{amsmath,amsfonts,bm}
\usepackage{dsfont}

\def\eqref#1{equation~\ref{#1}}

\def\1{\bm{1}}

\def\vc{{\bm{c}}}

\def\vt{{\bm{t}}}

\def\vv{{\bm{v}}}

\DeclareMathAlphabet{\mathsfit}{\encodingdefault}{\sfdefault}{m}{sl}
\SetMathAlphabet{\mathsfit}{bold}{\encodingdefault}{\sfdefault}{bx}{n}

\newcommand{\cv}[0]{{\boldsymbol{c}}}
\newcommand{\Cv}[0]{{\boldsymbol{C}}}

\newcommand{\tv}[0]{{\boldsymbol{t}}}
\newcommand{\Tv}[0]{{\boldsymbol{T}}}

\newcommand{\xmarkg}{\textcolor{lightgray}{\ding{55}}\xspace}%

\definecolor{cellcolor}{RGB}{163, 240, 247}

\usepackage[utf8]{inputenc}

\usepackage{listings}
\usepackage{xcolor}
\usepackage{amssymb}
\usepackage{pifont}
\lstdefinestyle{mystyle}{
    backgroundcolor=\color{backcolour},   
    commentstyle=\color{codegray},
    keywordstyle=\color{magenta},
    numberstyle=\tiny\color{codegray},
    stringstyle=\color{codepurple},
    basicstyle=\ttfamily\footnotesize,
    breakatwhitespace=false,         
    breaklines=true,                 
    captionpos=b,                    
    keepspaces=true,                 
    numbers=left,                    
    numbersep=5pt,                  
    showspaces=false,                
    showstringspaces=false,
    showtabs=false,                  
    tabsize=2,
    moredelim = [s][\color{purple}]{\\[}{\\]},
}

\newcommand\Tstrut{\rule{0pt}{2.6ex}}         
\newcommand\Bstrut{\rule[-0.9ex]{0pt}{0pt}}   
\usepackage{arydshln}
\newcommand*{\cmcsans}{\fontfamily{comic}\selectfont}
\DeclareTextFontCommand{\textcmcsans}{\cmcsans}
\usepackage{blindtext}
\usepackage{transparent}

\usepackage{float}
\usepackage[capitalize]{cleveref}
\crefname{section}{Sec.}{Secs.}
\Crefname{section}{Section}{Sections}
\Crefname{table}{Table}{Tables}
\crefname{table}{Tab.}{Tabs.}

\begin{document}

\title{Injecting Semantic Concepts into End-to-End Image Captioning}


\author{\large	  Zhiyuan Fang$^\spadesuit$, Jianfeng Wang$^\heartsuit$, Xiaowei Hu$^\heartsuit$, Lin Liang$^\heartsuit$, Zhe Gan$^\heartsuit$, \\ {\large Lijuan Wang$^\heartsuit$, {Yezhou Yang}$^\spadesuit$, {Zicheng Liu}$^\heartsuit$} \\  \ \ \ \ \  $^\spadesuit$Arizona State University, \ \ \ \ \ \ \ \ \ \ \ \ \ \ \ \ $^\heartsuit$Microsoft Corporation  \\
\texttt{\small \{zy.fang, yz.yang\}@asu.edu}\vspace{-.5mm}\\  \texttt{\small\{jianfw, xiaowei.hu, lliang, zhe.gan, lijuanw, zliu\}@microsoft.com}
}
\maketitle

\begin{abstract}
\vspace{-2mm}
Tremendous progresses have been made in recent years in developing better image captioning models, yet most of them rely on a separate object detector to extract regional features. Recent vision-language studies are shifting towards the detector-free trend by leveraging grid representations for more flexible model training and faster inference speed. However, such development is primarily focused on image understanding tasks, and remains less investigated for the caption generation task. In this paper, we are concerned with a better-performing detector-free image captioning model, and propose a pure vision transformer-based image captioning model, dubbed as \vitcap\!\!, in which grid representations are used without extracting the regional features. For improved performance, we introduce a novel Concept Token Network (CTN) to predict the semantic concepts and then incorporate them into the end-to-end captioning. In particular, the CTN is built on the basis of a vision transformer, and is designed to predict the concept tokens through a classification task, from which the rich semantic information contained greatly benefits the captioning task. Compared with the previous detector-based models, \vitcap drastically simplifies the architectures and at the same time achieves competitive performance on various challenging image captioning datasets. In particular, \vitcap reaches $138.1$ CIDEr scores on COCO-caption Karpathy-split, $93.8$ and $108.6$ CIDEr scores on nocaps and Google-CC captioning datasets, respectively.
\end{abstract}

\section{Introduction}
The task of image captioning aims to generate human-readable descriptive text from an image. Recent studies have witnessed its great development which are primarily reflected in the aspects of more advanced cross-modal fusion architectures~\cite{you2016image,vinyals2015show, xu2015show,rennie2017self,yang2019auto,cornia2020meshed,pan2020x,ting2019hierarchy,zhang2021rstnet}; more expressive object-centric features~\cite{anderson2018bottom,zhang2021multi} \& tags~\cite{li2020oscar,hu2020vivo,wang2020minivlm,fang2021compressing} obtained from a pre-trained object detection model; or learning general \textbf{\textit{V}}ision and \textbf{\textit{L}}anguage (VL) representations from large image-text corpus~\cite{zhou2020unified,xu2021e2e,li2020oscar,wang2021simvlm,wang2020minivlm,fang2021compressing}.

\begin{figure}[t!]
 \vspace{-3mm}
  \begin{center}
    \includegraphics[width=.46\textwidth]{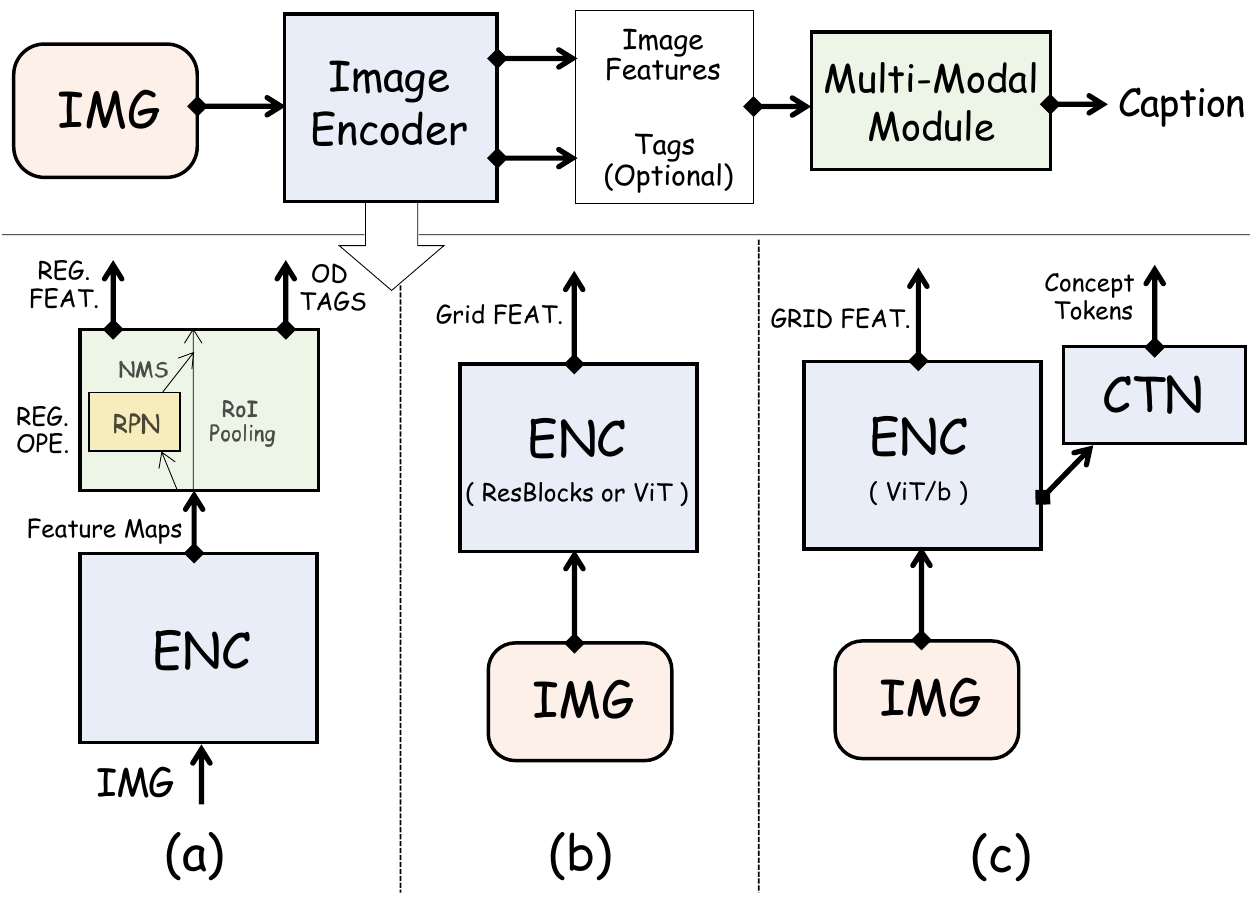} 
  \end{center}
  \vspace{-5mm}
    \caption{\small \textbf{Comparisons of different image captioning models}. Top: A general image captioning pipeline. Bottom: (a). Prevailing conventional models~\cite{li2019scale,zhang2021multi,hu2020vivo} which are based on an object detector to extract regional features. 
    Object tags~\cite{li2020oscar,zhang2021multi} can be optionally used to assist the text generation through a multi-modal decoder network. This usually requires regional operations (REG. OPE.) that are time consuming.
    (b). To eliminate the detection module, a ResNet variant~\cite{he2016deep} or Vision Transformer~\cite{kim2021vilt} can be applied as substitution to output the grid feature~\cite{xu2021e2e,wang2021simvlm}. 
    This replacement has been studied on the image understanding task recently but very few works
    focus on the generation task. 
    (c). Our proposed \vitcapp\!\!, which is detector-free and incorporates a novel Concept Token Network 
    to predict semantic concepts as tokens for the image captioning task.
    }
    \vspace{-2mm}
  \label{fig:abstract}
\end{figure}

Despite these significant advances, most of the mainstream captioning models~\cite{cornia2020meshed,pan2020x,ting2019hierarchy,zhang2021rstnet} rely heavily on a bulky object detector to provide regional visual representations for the multimodal interaction, as shown in Figure~\ref{fig:abstract}{\color{darkred}-a}. 
In spite of the superior performance brought by the object features, 
the ensuing difficulties occur as they: 1) \textbf{lead to heavy computational load} due to the regional operations (\ie, RPN, RoI Pooling, and NMS). These intermediate operations unavoidably cause training inefficiency and high inference latency at prediction stage~\cite{kim2021vilt,wang2020minivlm}; 2) \textbf{require box annotations} and largely limit the flexibility in training and application.  
To address these challenges, there is an emerging trend that more recent works propose to eliminate the detector for the VL pre-training in an end-to-end fashion~\cite{jiang2020defense,huang2020pixel,kim2021vilt,yan2021grid,wang2021simvlm}. In such detector-free design, a general visual encoder serves as a substitute for the detector and from which the grid features are produced for later cross-modal fusion, as in Figure~\ref{fig:abstract}{\color{darkred}-b}.
Heretofore, the majority of these works mainly focus on the image understanding task, which is typically cast as a classification problem, and only a few of them shed light on the generation task.
In~\cite{xu2021e2e}, the image is encoded with ResNet~\cite{he2016deep} and 
the performance ($117.3$ CIDEr on COCO~\cite{xu2021e2e})
is still far from the state-of-the-art detector-based approach ($129.3$ CIDEr with VinVL-base~\cite{zhang2021multi}).
The challenge remains uncharted and insufficiently investigated regarding \textit{how to build a stronger detector-free image captioning model.}

Previous efforts~\cite{li2020oscar,hu2020vivo,zhang2021multi,wang2020minivlm,fang2021compressing} have demonstrated that the object tags play an important role in improving the captioning performance. 
Instead of gleaning the object tags from the detector, we introduce a novel fully \textbf{VI}sion \textbf{T}ransformer based image \textbf{CAP}tioning model, dubbed ViTCAP, with a lightweight Concept Token Network (CTN) that produces concept tokens (see Figure~\ref{fig:abstract}{\color{darkred}-c}). 
ViTCAP is constructed on the basis of a vision transformer~\cite{dosovitskiy2020image} as the stem image encoder. Our vision transformer backbone starts with encoding the image and produces grid features, on top of which the CTN branch is then applied to predict semantic concepts of images. We represent the semantic concepts at the token level instead of the tag level to avoid the tokenization. 
The multi-modal module then takes the input of both grid representations and Top-$K$ concept tokens for decoding. During training, the CTN is optimized to predict the pseudo ground-truth concepts extracted from image captions via a simple classification task. We also investigate to adopt the object tags from the detector as the pseudo ground-truth, and empirically observe no further improvement. Overall, this straight-forward design allows the injection of semantic concepts into the multi-modal fusion module with abundant semantics, and is critical for the improved captioning performance.

Our ablative analysis suggests that, with no bells and whistles, simple vanilla transformer architecture based \vitcap 1) significantly outperforms existing detector-free captioning models;
2) surpasses most detector-based models and 3) approaches the state-of-the-art detector-based models. 
In particular, \vitcap achieves $138.1$ CIDEr scores on COCO-caption Karpathy split~\cite{lin2014microsoft}, $108.6$ on Google-CC~\cite{sharma2018conceptual}, and $95.4$ on nocaps~\cite{agrawal2019nocaps} datasets.

\noindent To summarize our contributions:
\begin{itemize} [leftmargin=8pt]
    \vspace{-2mm}
    \item We present a detector-free image captioning model  \vitcap\!\! with fully transformer architecture, where it leverages grid representations without regional operations. 
    \vspace{-2mm}
    \item We propose to inject semantic concepts into end-to-end captioning by learning from open-form captions. We find that our proposed concept classification training and concept tokens significantly benefit the captioning task.
    \vspace{-2mm}
    \item Extensive evaluations on multiple captioning datasets confirm the validity of our method. \vitcap achieves competitive or even leading results amongst detector-based prior arts with clear inference-time advantages.
\end{itemize}

\section{Related Work}
\noindent \textbf{Image Captioning} aims to produce an open-form and human-readable textual description that summarizes the content of an image. Most previous captioning models unanimously~\cite{anderson2018bottom,you2016image,vinyals2015show, xu2015show,rennie2017self,ting2019hierarchy, gan2017semantic, hu2020vivo, fang2020video2commonsense} use detector based visual encoder like Faster-RCNN~\cite{ren2015faster} to extract visual features, and apply decoders  like RNN, LSTM or Transformer for caption generation. Existing efforts on image captioning are reflected from the perspective of novel architectures~\cite{pan2020x,cornia2020meshed,zhang2021rstnet}, more effective learning objectives~\cite{rennie2017self,luo2018discriminability,hu2020vivo}, or large-scale VL pre-training~\cite{zhou2020unified,li2020oscar,zhang2021multi}, \etc. Some recent works~\cite{anderson2018bottom,zhang2021rstnet} arrive at an empirical conclusion that a strong object detector is necessary, providing clean and unambiguous regional features for objects. Li \etal~\cite{li2020oscar,hu2020vivo} show that object tags output from the detectors play a critical role as anchoring points in VL tasks across modalities. Following this,~\cite{zhang2021multi} proposes to adopt a strengthened detector to obtain regional features and expanded object tags covering both entities and attributes for VL tasks. Nevertheless, object detectors hinder the VL models to be deployed on edge devices, known for their snail's pace at inference.  

\begin{figure*}[t!]
  \begin{center}
    \includegraphics[width=\textwidth, height=0.35\textwidth]{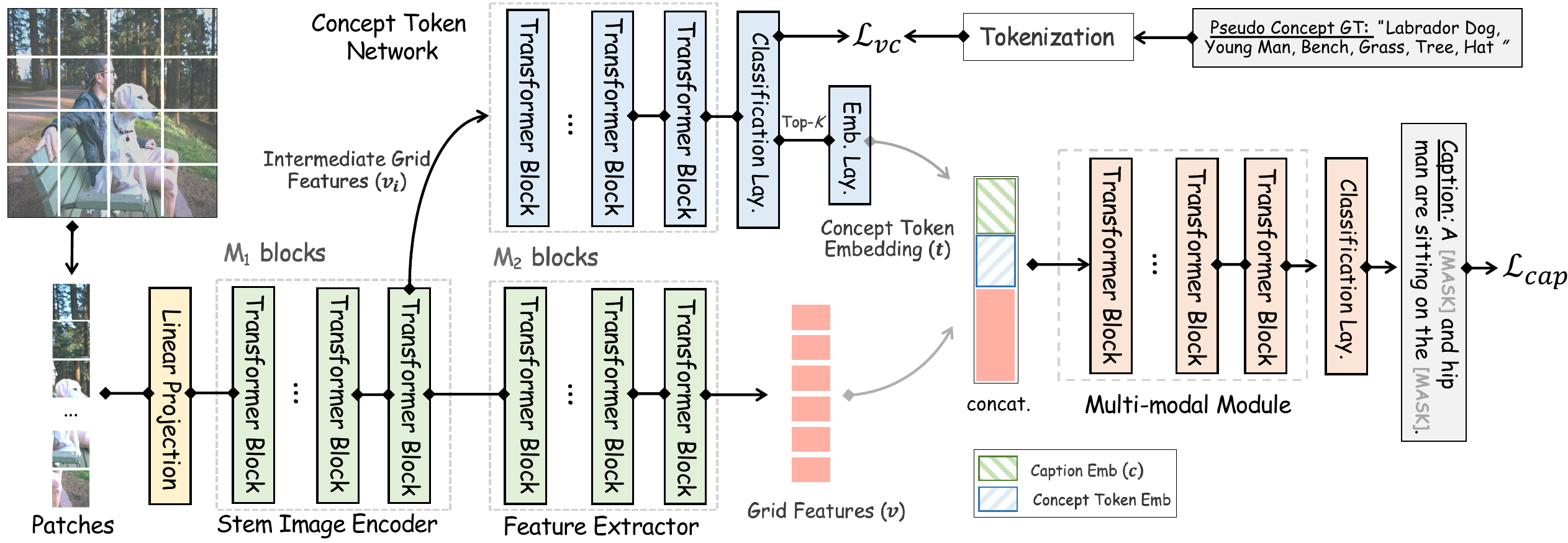}
    \vspace{-8mm}
  \end{center}
    \caption{\small \textbf{Architecture of our proposed \vitcapp image captioning model}. \vitcapp is a detector-free image captioning model based on the vision transformer, where image patches are encoded into continuous embeddings as grid representations. The CTN branch roots from an intermediate block of the image encoder, and is a shallow transformer architecture (\eg, 4 self-attention blocks). The CTN is trained via a classification task using object tags gleaned from the Teacher VLM's detector as pseudo-labels and the keywords parsed from image captions as the semantic concept ground-truth. During captioning, the CTN-produced concept tokens from the semantic concept vocabulary are then concatenated with the grid representations and fed into the multi-modal module for decoding. Best viewed in color.
    }
  \label{fig:architecture}
  \vspace{-3mm}
\end{figure*}

\vspace{1mm}
\noindent \textbf{Efficient VL Models}. Several recent efforts build efficient VL Models that either optimize the object detector for feature extraction with faster inference speed, or adopt non-detector image encoders. For instance, MiniVLM~\cite{wang2020minivlm} first proposes an EfficientNet~\cite{tan2019efficientnet} based lightweight detector.~\cite{jiang2020defense} revisits grid features for VQA task with great performance and fast inference speed.~\cite{huang2020pixel,kim2021vilt,xu2021e2e,wang2021simvlm,dou2021empirical} also inherit such detector-free design and use architecture like ResBlocks~\cite{he2016deep} for image encoding. On the other side, DistillVLM~\cite{fang2021compressing} introduces VL distillation that facilities VL pre-training \& fine-tuning for small transformer architectures;~\cite{gan2021playing} proposes to prune the transformer architecture and shows that close performance can be maintained at $50$\%-$70$\% model sparsity.

\section{ViTCAP}
Existing image captioning models usually consist of an object detector module (\textbf{Detector}) to extract regional feature ($\vv^{T}$) from the raw image (\,$\mathbf{I}$\,), and a multi-modal module (\textbf{MM}) to generate a textual description (\,$\vc$\,). Several recent works~\cite{li2020oscar,zhang2021multi} show that the object tags ($\vt^{T}$) extracted from the detector can serve as anchoring points across modalities, and are essential for various VL tasks. This procedure can be expressed as follows:
\begin{equation}
    (\vv^{T}, \vt^{T}) = \text{\textbf{Detector}}(\, \mathbf{I}\,), \ \ \ \cv = \text{\textbf{MM}}(\vv^{T}, \tv^{T}).
\end{equation}
Several VL models~\cite{huang2020pixel,kim2021vilt,xu2021e2e} obtain a great improvement in inference speed by using general image encoders without regional operations. However, these models are unable to utilize the image tags due to the absence of a detector.  

In this work, we aim to build a detector-free captioning model with concept tokens containing rich semantics, coming from a novel Concept Token Network (CTN).
An overview of \vitcap is depicted in Figure~\ref{fig:architecture}. 
The raw image is firstly fed into the image encoder to generate the intermediate representations ($\vv_{i}$) and the final grid representations ($\vv$).
A CTN branch then takes $\vv_{i}$ as the input and predicts concept tokens ($\tv$), followed by the multi-modal module that allows the interactions across modalities and generates caption ($\cv$). We adopt the fully transformer~\cite{vaswani2017attention} framework in all modules, but the image encoder and CTN modules are not architecture-specific. The overall pipeline can be summarized as:
\begin{equation}
    (\!\vv_{i}, \vv\!) \!=\! \text{\textbf{Encoder}}(\, \mathbf{I}\,), \ \ \tv \! = \! \text{\textbf{CTN}}(\vv_{i}), \ \ \cv \! = \! \text{\textbf{MM}}(\vv, \tv).
\end{equation}
In the following, we first introduce how the vision transformer produces grid representations and our proposed CTN in Section~\ref{sec:tagger}, and the overall training losses in Section~\ref{sec:training}.

\subsection{Model Structure}
\noindent \textbf{Vision Transformer.}
\label{sec:vit}   
The transformer architecture and its instantiations (\eg, BERT~\cite{devlin2018bert}, GPT~\cite{brown2020language}) are well-known for their remarkable performances on natural language processing tasks, which are mostly attributed to the self-attention design. Recent efforts have advanced this to vision tasks, \ie, Vision Transformer (ViT)~\cite{dosovitskiy2020image}. 
We use ViT as the backbone of the image encoder to produce grid representations ($\vv_i$ and $\vv$ ). 
To be specific, the raw image $\mathbf{I}\in\mathbb{R}^{H\times W \times3}$ is partitioned into $N$ disjoint patches. The size of each patch is $P\!\times\!P\!\times\!3$ and the number of patches $N\!$ is 
${(H W)}/{P^2}$. 
These patches are then flattened and projected into patch embedding of dimension $d$ via a trainable linear projection layer. Concatenated with a special \texttt{[CLS]} token, these patch representations are added with learnable positional embeddings and then sent into $M$ consecutive transformer blocks thereafter. 
To this end, we use the final representation as the grid features $\vv$, and extract the output of the first $M_1$ blocks as the intermediate representations $\vv_i$, which is the input of the Concept Token Network for concept predictions as detailed below. 

\vspace{1mm}
\noindent \textbf{Concept Token Network.}
\label{sec:tagger}
The Concept Token Network (CTN) is composed of $M_2$ transformer blocks to process the intermediate
features $v_i$. The output representation corresponding to \texttt{[CLS]} is used to predict the 
concept token with a multi-linear perceptual (MLP) network. 
The vocabulary of the concept token is identical with the one used for the captions. 
It is noted that we predict the concept in the token level rather than in the tag level, and thus the top-$K$ ($K=50$ in our experiments) tokens can be directly used by the multi-modal decoding module for auto-regressive decoding. In~\cite{li2020oscar,zhang2021multi}, the object tags are predicted from the object detector, while we eliminate the detection module to remove the dependency of the box annotations. Another difference lies in the tag/concept vocabulary. The existing approaches apply the tag list from the dataset as the vocabulary which are pre-defined and need an extra tokenization operation. Instead, our concept token vocabulary is shared with the one for captions and also removes the tokenization step.




\vspace{1mm}
\noindent \textbf{Multi-Modal Fusion Module.}
\label{sec:multimodal}
Our multi-modal fusion module is a shallow network composed of multiple transformer blocks, and we follow~\cite{radford2018improving,brown2020language} to apply the \texttt{seq2seq} attention mask to generate the caption token in an auto-regressive way.
First, the Top-$K$ concept tokens' indices are mapped to token embeddings through an embedding layer $l_c$.
Then, the module takes as input the concatenation of concept token embeddings ($\tv$) and grid representations ($\vv$) to generate the description, where we append a mask token \texttt{[MASK]} to the previous generated tokens (empty at very beginning) to predict the next token one by one.
With the \texttt{seq2seq} attention mask, the generated token (including the appended \texttt{[MASK]} token) is able to access the preceding tokens and $(\tv, \vv)$, while $(\tv, \vv)$ has no access to the generated tokens.  
The generated caption token is also mapped through an embedding layer $l_d$.
In experiments, we make the two embedding layers ($l_c$ and $l_d$) shared to reduce the parameter size as the result is similar to two separate layers (see Appendix for results). 

\subsection{Model Training}
\label{sec:training}
The training of \vitcap is composed of the CTN and the captioning training.

CTN is used to predict the image concepts. However, the widely-used VL pre-training dataset contains only the image descriptions without the tags. To address the issue, one can simply retrieve the concepts from the open-form captions (\eg, by extracting nouns or adjective words as keywords) as the pseudo ground-truth concepts, or alternatively leverage a pre-trained object detector (\eg on Visual Genome~\cite{krishna2017visual}) to produce the image tags (remove the bounding boxes). Empirically, we observe that by using caption extracted concepts lead to better results.
We optimize the CTN to predict the target concepts via a multi-label classification task. Due to the extremely imbalanced semantic concepts distribution (certain concepts appear much frequently than the rest), we adopt the simplified asymmetric focal loss~\cite{ben2020asymmetric,liu2021query2label,lin2014microsoft} which shows great performances handling sample imbalance problems for the multi-label classification task. The overall concept classification loss can be expressed as: 
\begin{align}
\mathcal{L}_{{vc}} = \mathbb{E}_{\vv_i\sim D}f_{\theta}(p\ |\ \vv_i),
\end{align}
\begin{equation}
      f_{\theta}(p\ |\ \vv_i) = \frac{1}{K} \sum^{K}_{k=1}
\begin{cases}
    (1-p_{k})^{\gamma_{+}}\cdot\text{log}(p_k), & +, \\ p_{k}^{\gamma_{-}} \cdot \text{log}(1-p_k), & -,
\end{cases} 
\end{equation}
$p_k \in [0, 1]$ denotes the output probability for the $k$-th class and ${\pm}$ specifies whether the class is the pseudo ground-truth concept. Despite the rarity of positive samples, setting parameters $\gamma_{+} < \gamma_{-}$ decouples its decay rates from the deluge of negative samples and emphasizes more the contribution of the positive. We set parameters $\gamma_{+}=0$ and $\gamma_{}-=1$ as~\cite{liu2021query2label} in our experiment.


For the captioning training, the multi-modal module takes the Caption-Concept Token-Feature triple $(\cv, \tv, \vv)$ as input, 
where $\cv = \{\cv_1, \dots \cv_T\}$ are the masked input words after tokenization and we set the mask probability $=15\%$. The masked tokens are replaced with the special token \texttt{[MASK]}.
The prediction of masked token at the position $t$ is conditioned on the preceding tokens ($\cv_{<t}$), visual representations ($\vv$) and the concept tokens ($\tv$). We train our model parameters $\theta$ by minimizing the negative log-likelihood over the masked tokens:
\begin{equation}
    \mathcal{L}_{cap} = -\mathbb{E}_{\Tv\sim D}\Big[\text{log}\!\!\!\!\prod_{\hat{\cv_{t}}\sim \Cv_{M}}\!\!\!\! P_{\theta}(\hat{\cv_{t}}|\cv_{<t}, \tv, \vv)\Big],
\end{equation}
where $\Cv_{M}$ refers to the ground-truth set of the masked tokens.

Recent works~\cite{fang2021compressing,liu2021kd} reveal that by leveraging the knowledge distillation technique~\cite{hinton2015distilling}, the VL model can be improved compared to the non-distilled counterpart using a pre-trained Teacher VL model. In our training, we experiment with applying a trained detector-based captioning model as the Teacher (parameterized by $\theta_{t}$), \ie, VinVL~\cite{zhang2021multi}, to assist the training of \vitcap\!\!. Note that the Teacher model is a two-stage VL model adopting regional features and object tags from the detector, yielding discrepant visual features with \vitcap\!\!, and hence the distillation objectives like attention-map loss and hidden-states loss are not directly applicable as in~\cite{fang2021compressing}. We adopt the classification distillation loss over the masked token probabilities between the predictions from the Student ($P_\theta$) and Teacher ($P_{\theta_t}$) models:
\begin{equation}
    \mathcal{L}_{dis} = \mathbb{E}_{\Tv\sim D}\Big[\sum_{\hat{\cv_t}\sim\Cv_{M}}\!\!\!\text{KL}{\Big(}P_\theta(\hat{\cv_t}), P_{\theta_t}(\hat{\cv_t}){\Big)}\Big],
\end{equation}
where KL$( \ , \ )$ is the Kullback–Leibler divergence.
Overall, our final loss is then the combination of the terms:
\begin{equation}
    \mathcal{L} = \mathcal{L}_{vc} + \mathcal{L}_{cap} + \mathcal{L}_{dis}.
\end{equation}

\section{Experiment}
We now introduce the implementation details of \vitcap and empirically verify the validity of our proposed training schema from different aspects. 
To highlight the generalizability of \vitcap\!\!, we benchmark performances of \vitcap and compare it with prior arts on multiple image captioning testbeds. We then exhaustively study the effect of our proposed concept tokens, the benefits of pre-training at scale, the effect of VL distillation, \etc. In the end, we visualize the attention maps of \vitcap and provide in-depth discussion.

\subsection{Datasets}

\vspace{1mm}
\noindent \textbf{Pre-training Datasets.}
In our experiment, we aggregate image-text pairs from Google-CC~\cite{sharma2018conceptual}, SBU Caption dataset~\cite{ordonez2011im2text}, MS COCO~\cite{lin2014microsoft} and Visual Genome dataset~\cite{krishna2017visual} to form the pre-training corpus.  In total, our pre-training corpus contains $9.9$M image-text pairs and $4.1$M independent images, and we follow~\cite{lu202012} to de-duplicate testing images exist in evaluating datasets. Details of the pre-training corpus can be found in the Appendix. 

\vspace{1mm}
\noindent \textbf{Evaluation Datasets.}
We report performances of \vitcap on {{COCO captions}} (Karpathy split)~\cite{lin2014microsoft}, {{Google-CC}}~\cite{sharma2018conceptual}, and {{nocaps}}~\cite{agrawal2019nocaps} datasets. We follow Karpathy’s split and use~$113$k, $5$k and $5$k images for training, validation and testing respectively on MS COCO dataset. As regards to Google-CC, we follow~\cite{sharma2018conceptual} and use its training split containing~$3$M image-text pairs for training, and report the performances on validation split with~$16$K image-text pairs. To test the generalization of \vitcap\!\!, we also report the performances on nocaps dataset~\cite{agrawal2019nocaps}, a benchmark consisting of $166$k human-generated captions describing $15$k images in the wild collected from the OpenImages dataset~\cite{shao2019objects365}.

\subsection{Implementation Details}
\vspace{1mm}
\noindent \textbf{Architecture.}
Our \vitcap is based on a Vision Transformer base (ViT/b) architecture consisting of $M=12$ consecutive transformer blocks, with hidden size as $768$, and $12$ attention heads. In our experiment, we set the patch size as $16 \times 16$ and resize the shorter side of the image to $384$. We use $M_1=8$ transformer blocks in Stem Image Encoder to extract the intermediate grid representations and use $M_2=4$ transformer blocks for the CTN branch. When enlarging the size of CTN and Feature Extractor to $M_2=12$ transformer blocks, it is equivalent to two independent networks for the computation of Concept Token/Embedding and Grid Feature respectively. We adopt this design with more learnable parameters in our ViTCAP with large scale pre-training (see ViTCAP$^*$ in Table~\ref{tab:COCO}).
Data augmentations are applied on raw images before the linear projection as~\cite{dosovitskiy2020image} including \textit{ColorJitter}, \textit{horizontal flipping}, \etc.

\vspace{1mm}
\noindent \textbf{Two-stage Training.}
Training both the CTN branch jointly with the captioning task jointly from scratch is challenging, we observe that using a pre-trained CTN with stable and consistent concept prediction throughout the training leads to superior captioning results. Thus in practice, we first conduct concept classification training for a good concept prediction, and then train the model with both tasks. 
Such strategy prevents the ``\textit{cold-start}'' issue when the initially produced concepts are mostly random, impairing the captioning training. During the joint captioning \& concept branch training, we reduce the learning rate for both the Stem Image Encoder and CTN branch by a factor of $\alpha$ ($\alpha = 10$) and keep the predicted concepts relatively consistent but still slowly adapted throughout the training. 
\begin{itemize} [leftmargin=8pt]
    \item \textbf{Concept Classification.} The concept classification is conducted on an aggregated dataset with $4.1$M images (see later section for details). To obtain the pseudo ground-truth concepts, we experiment with using the NLTK~\cite{loper2002nltk} toolkit to parse out the nouns and adjectives as the target concepts, or simply use all tokens in captions as targets for the classification task. For the detector-produced tags, we take advantage of a ResNeXt-$152$ C4 architecture based object-attribute detector that has been well-trained~\cite{zhang2021multi} to produce image tags as pseudo-labels for concept classification training. We only retain image tags with confidence score $> 0.2$ from the detector and acquire $50$ tags at most per image. For classification training, the model is initialized from the ImageNet-$21$k~\cite{krizhevsky2012imagenet} pre-trained checkpoint\footnote{\url{https://github.com/lucidrains/vit-pytorch}.}, and is optimized for $10$ epochs using AdamW~\cite{you2019large,reddi2019convergence} optimizer. The batch size is $1,024$. The initial learning rate is $5e-5$ and is linearly decayed to $0$. 
    \item  \textbf{Captioning Training.} For the joint optimization, we apply the well-trained model after concept classification to initialize Stem Image Encoder, CTN and the feature extractor. The initial weights in the feature extractor are copied from the CTN branch, as the architecture for grid feature extractor is the same as the CTN branch. We set base learning rate~$lr=1e-4$, batch-size $=512$ and train the model for $30$ epochs using AdamW optimizer, and set weight decay$=0.05$. 
\end{itemize}
\noindent \textbf{Evaluation.} We evaluate the quality of the generated captions using the prevailing metrics including BLEU@$4$~\cite{papineni2002bleu}, METEOR~\cite{banerjee2005meteor}, CIDEr~\cite{vedantam2015cider}, ROUGE~\cite{lin2004rouge} and SPICE~\cite{anderson2016spice}. During inference, we use beam search (beam size $= 1$) for decoding. There exist many evaluating metrics studying the qualities of the generated captions, including Self-CIDEr~\cite{wang2020diversity}, SMURF~\cite{feinglass2021smurf} and from different aspects~\cite{jiang2019tiger,wang2021faier,hessel2021clipscore}. In the Appendix, we conduct more studies studying the diction quality of our generations using SMURF~\cite{feinglass2021smurf} metric.

\begin{table*}[t] 
\small
\centering
\renewcommand{\arraystretch}{1.1} 
\setlength\tabcolsep{4.5pt}
\scalebox{1.0}{
\begin{tabular}{p{24mm} c p{9mm}  c c c c  c c c c c c}
\toprule
 \\[-2.5ex]
\multirow{2}{*}{\textbf{Methods}} & \multirow{2}{*}{V. ENC.} & \multirow{2}{*}{\# \textit{I-T}} & \multicolumn{5}{c}{{\texttt{\textbf{Cross-Entropy Loss}}}} & \multicolumn{5}{c}{{\texttt{\textbf{CIDEr Optimization}}}} \\
\cmidrule(r){4-8} \cmidrule(l){9-13}
&  &  & \multicolumn{1}{c}{B@4 } & M    & R    & C     & \multicolumn{1}{c}{S}  &  B@4  & M  & R  & C     & S  \\	\hline \Tstrut\Bstrut 
{$^{{\color{darkred}{\text{\textbf{Detector}}}}\text{ w.o. }\color{darkred}{\text{\textbf{VLP}}}}$} & & & & & & & & & \\[-6pt]
{\cellcolor{blue!3} RFNet~\cite{jiang2018recurrent}}       & Ensemble & \ \ \xmarkg & $35.8$ & $27.4$ & $56.5$ & $112.5$ & $20.5$ & $36.5$ & $27.7$ & $57.3$ & $121.9$ & $21.2$ \\
{\cellcolor{blue!3}BUTD~\cite{anderson2018bottom}}    & F-RCNN$_{101}$ & \ \ \xmarkg & $36.2$ & $27.0$ & $56.4$ & $113.5$ & $20.3$ & $36.3$ & $27.7$ & $56.9$ & $120.1$ & $21.4$ \\
{\cellcolor{blue!3}LBPF~\cite{yao2018exploring}}        & F-RCNN$_{101}$ & \ \ \xmarkg & 37.4 & 28.1 & 57.5 & 116.4 & 21.2 & 38.3 & 28.5 & 58.4 & 127.6 & 22.0 \\
{\cellcolor{blue!3}SGAE~\cite{yang2019auto}}        & F-RCNN$_{101}$ & \ \ \xmarkg & $36.9$ & $27.7$ & $57.2$ & $116.7$ & $20.9$ & $38.4$ & $28.4$ & $58.6$ & $127.8$ & $22.1$ \\
{\cellcolor{blue!3}AoANet~\cite{huang2019attention}}      & F-RCNN$_{101}$ & \ \ \xmarkg & $37.2$ & $28.4$ & $57.5$ & $119.8$ & $21.3$ & $38.9$ & $29.2$ & $58.8$ & $129.8$ & $22.4$ \\
{\cellcolor{blue!3}M$^{2}$ Transfm.~\cite{cornia2020meshed}} & F-RCNN$_{101}$ & \ \ \xmarkg & - & - & - & - & - & $39.1$ & $29.2$ & $58.6$ & $131.2$ & $22.6$\\ 
{\cellcolor{blue!3}X-LAN~\cite{pan2020x}}       & F-RCNN$_{101}$ & \ \ \xmarkg & $38.2$ & $28.8$ & $58.0$ & $122.0$ & $21.9$  & $39.5$ & $29.5$ & $59.2$ & $132.0$ & $23.4$ \\
{\cellcolor{blue!3}RSTNet~\cite{zhang2021rstnet}} & RESNeXt$_{152}$ & \ \ \xmarkg & - & - & - & - & - & $40.1$ & $29.8$ & $59.5$ & $135.6$ & $23.3$\\ 
\hline \Tstrut\Bstrut 
{$^{{\color{darkred}{\text{\textbf{Detector-Free}}}}\text{ w.o. }\color{darkred}{\text{\textbf{VLP}}}}$} & & & & & & & & & \\[-6pt]
{\cellcolor{orange!3}ViTCAP \ \ \ (Ours)}    & ViT$_{b}$  & \ \ \ \xmarkg & $35.7$ & $28.8$  & $57.6$ & $121.8$ & $22.1$ & $40.1$ & $29.4$ & $59.4$ & $133.1$ & $23.0$ \\
\hline \Tstrut\Bstrut  
{\cellcolor{red!3}$^{{\color{darkred}{\text{\textbf{Detector}}}}\text{ w. }\color{darkred}{\text{\textbf{VLP}}}}$}  &  &  &  &  &  &  & & & & \\[-6pt]
{\cellcolor{red!3} UVLP~\cite{zhou2020unified}}        & F-RCNN$_{101}$ & \ $4$M & $36.5$ & $28.4$ & - & $116.9$ & $21.2$ & $39.5$ & $29.3$ & - & $129.3$ & $23.2$ \\
{\cellcolor{red!3} MiniVLM~\cite{wang2020minivlm}}      & Eff-DET & $14$M & $35.6$ & $28.6$ & - & $119.8$ & $21.6$ & $39.2$ & $29.7$ & - & $131.7$ & $23.5$ \\
{\cellcolor{red!3}DistillVLM~\cite{fang2021compressing}}   & Eff-DET & \ $7$M & $35.6$ & $28.7$ & - & $120.8$ & $22.1$ & - & - & - & - & - \\
{\cellcolor{red!3} OSCAR$_\text{b}$~\cite{li2020oscar}}        & F-RCNN$_{101}$ & \ $7$M & $36.5$ & $30.3$ & - & $123.7$ & $23.1$ & $40.5$ & $29.7$ & - & $137.6$ & $22.8$ \\
{\cellcolor{red!3}UNIMO$_\text{b}$~\cite{li2020unimo}} & F-RCNN$_{101}$ & \ $9$M & $38.8$ & - & - & 124.4  & - & - & - & - & - & -\\ 
{\cellcolor{red!3}VL-T5~\cite{cho2021unifying}} & F-RCNN$_{101}$ & \ $9$M & - & - & - &  $116.5$ & - & - & - & - & - & -\\
{\cellcolor{red!3}VinVL$_\text{b}$~\cite{zhang2021multi}}        & RESNeXt$_{152}$ & \ $9$M  & $38.2$ & $30.3$ & - & $129.3$ & $23.6$ & \textbf{$40.9$} & \textbf{$30.9$} & - & \textbf{$140.4$} & \textbf{$25.1$} \\
\hline \Tstrut\Bstrut 
{$^{{\color{darkred}{\text{\textbf{Detector-Free}}}}\text{ w. }\color{darkred}{\text{\textbf{VLP}}}}$} & & & & & & & & & & \\[-6pt]
{\cellcolor{orange!3}ViLT-CAP$^{\color{black}{\ \spadesuit}}$}  & ViT$_{b}$ & \ $10$M & $33.7$ & $27.7$ & $56.1$ & $113.5$ & $20.9$ & - & - & - & - & - \\
{\cellcolor{orange!3}E2E-VLP~\cite{xu2021e2e}} & ResNet$_{50}$ & \ \ $6$M & $36.2$ & - & - & $117.3$ & - & - & - & - & - & - \\
{\cellcolor{orange!3}ViTCAP$^*$\ \ \ (Ours)} & ViT$_{b}$  & \ $10$M & $\textbf{36.3}$ & $\textbf{29.3}$ & $\textbf{58.1}$ & $\textbf{125.2}$ & $\textbf{22.6}$ & $\textbf{41.2}$ & $\textbf{30.1}$ & $\textbf{60.1}$ & $\textbf{138.1}$ & $\textbf{24.1}$\\
\bottomrule
\end{tabular}
}
\vspace{2mm}
\caption{\small Performance comparisons on \coco Karpathy split~\cite{karpathy}, where B@$4$, M, R, C denote BLEU@$4$, METEOR, ROUGE-L, CIDEr and SPICE scores. All values are reported as percentages (\%). We compare the \vitcap with previous state-of-the-art detector-based baselines (without the VLP) in the first section, and detector-based baselines (with large scale pre-training) in the third section, and the detector-free methods with pre-training in the last section.
V. ENC. denotes visual encoders for feature extraction; \# \textit{I-T} refers to the number of image-text pairs used in pre-training (in millions). ViTCAP$^*$ is a larger version of ViTCAP with more parameters.
$^{\color{black}{\spadesuit}}$ is the results we achieved using the ViLT~\cite{kim2021vilt} pre-trained checkpoint for image captioning task (see Appendix for more explanation). 
}
\label{tab:COCO}
\vspace{-1mm}
\end{table*}

\subsection{Main Results}
We perform extensive comparisons of \vitcap with the prior arts. Table~\ref{tab:COCO} presents the captioning results on MS COCO dataset where the models are trained with cross-entropy loss or optimized with CIDEr as reward~\cite{rennie2017self}. 
We compare \vitcap with 1). ``\textit{detector w/o VLP}'' models with complex architectural modifications. These models ~\cite{huang2019attention,cornia2020meshed,pan2020x,zhang2021rstnet} all come unanimously with heavy computational burdens and extra learnable parameters. 2). ``\textit{detector w. VLP}'': prevailing detector-based VL models pre-trained with a large VL corpus and then fine-tuned on image captioning tasks. 3). ``\textit{detector-free}'' methods: the end-to-end trainable image captioning models without object detector (with or without pre-training).

\noindent \textbf{Without VLP.} To compare fairly with the detector-based baselines without VLP,  we adopt the VinVL tags as concept sources instead of the captions to guarantee that \emph{no additional captions have been exploited} during the concept classification training. Note that the knowledge distillation objective is not applied for this experiment as it introduces extra knowledge from the pre-training of Teacher model.
On COCO-caption Karpathy split, our \vitcap achieves similar results and even surpasses most existing detector-based methods, \ie, CIDEr score $121.8$, using caption extracted concepts. It is worth mentioning that the architectures of most existing detector-based methods are deliberately designed, \eg, the self-attention module in X-LAN~\cite{pan2020x} has 2$^{nd}$ interactions for multi-modal inputs, M$^{2}$ Transformer~\cite{cornia2020meshed} has the multi-level representation of the relationships between image regions, \etc. \vitcap adopts the simplest vanilla transformer architecture without any bells and whistles. This proves the effectiveness of our proposed learning paradigm. 
The ablations in the later section comprehensively explore the benefits of CTN and the knowledge distillation technique.

\noindent \textbf{With VLP.}
We observe a clear performance gain of \vitcap after the large scale pre-training ($3.0$ higher CIDEr scores), better than most detector-based VL methods: \eg, $125.2$ \vs $123.7$ (OSCAR$_{b}$), and $0.8$ higher than UNIMO$_{b}$, $8.7$ higher than VL-T$5$ when pre-trained on similar VL corpus. This conclusion is further supported by results of other metrics. \vitcap approaches the state of the art, only $2.3$ lower than VinVL in CIDEr scores after CIDEr optimization, considering the fact that VinVL used ResNeXt$_{152}$-based object detector.
Compared with detector-free baselines, \vitcap outperforms all existing works with an obvious discrepancy: $11.7$ CIDEr scores higher than the ViLT-CAP~\cite{kim2021vilt} and $7.9$ higher than E2E-VLP~\cite{xu2021e2e}. 

\subsection{Ablative Study}
We now comprehensively study \vitcap\!'s performance gain from different aspects, \ie, knowledge distillation, the effect of concept tokens, and large-scale pre-training.

\begin{table}[t]
\centering
\renewcommand{\arraystretch}{1.15} 
\setlength\tabcolsep{4.5pt}
\scalebox{1.0}
{
\small
\begin{tabular}{p{22mm} | c c c c c }
\toprule
\multirow{2}{*}{\textbf{Concept Source}} & \multicolumn{4}{c}{{ \texttt{\textbf{\ \ \ \ \ COCO Captioning}}}}  \\ 
 & {B@4} & {\ \ M} & {\ \ R} & {\ \ C} & {\ \ S} \\
\hline
\cellcolor{red!3}\xmarkg  & \cellcolor{gray!5}$33.9$ & $27.8$ & $56.4$ & \cellcolor{gray!5}$114.8$ & $21.3$ \\
\cellcolor{red!3}BUTD~\cite{anderson2018bottom}  & \cellcolor{gray!15}$35.0$ & $28.2$ & $56.9$ & \cellcolor{gray!25}$117.4$& $21.3$ \\
\cellcolor{red!3}VinVL~\cite{zhang2021multi}  & \cellcolor{gray!25}$35.6$ & $28.6$ & $57.4$ & \cellcolor{gray!35}$119.7$ & $21.8$ \\
\cellcolor{red!3}CAPTION & \cellcolor{gray!35}$35.6$ & $28.7$  & $57.6$ & \cellcolor{gray!35}$120.9$ & $21.8$ \\
\cellcolor{red!3}VinVL $\rightarrow$ CAP.$^{\spadesuit}$& \cellcolor{gray!35}$35.9$ & $28.6$  & $57.6$ & \cellcolor{gray!40}$121.3$ & $21.9$ \\
\cellcolor{red!3} CAPTION$^{\spadesuit}$ & \cellcolor{gray!45}$35.7$ & $28.8$ & $57.6$ & \cellcolor{gray!45}$121.8$ & $22.1$ \\ 
\bottomrule
\end{tabular}
}
\vspace{2mm}
\caption {\small Adopting various sources of semantic concept leads to different performances. ``CAPTION'' represents the baseline extracting keywords from open-form captions; ``$^{\spadesuit}$'' is the baseline using all words in captions as target concepts; ``BUTD'' and ``VinVL'' represent using the object tags produced by the object detector from~\cite{anderson2018bottom} and~\cite{zhang2021multi} as target semantic concepts, respectively. ``VinVL $\rightarrow$ CAP.'' represents adopting detector tags~\cite{zhang2021multi} during first stage of concept classification and using caption extracted tags during the second stage.}
\label{tab:tags}
\vspace{-4mm}
\end{table}

\noindent \textbf{Semantic Concept Sources.} We study the effects of different semantic concept sources, \ie, from object detectors~\cite{anderson2018bottom,zhang2021multi}, captions-extacted concepts, and the combination of them. Table~\ref{tab:tags} lists the performances of \vitcap on the COCO caption dataset with various semantic concepts sources. 
Open-form captions are the most accessible source to directly obtain semantic concepts, although these descriptions can sometimes be noisy, inaccurate and incomplete. ``CAPTION'' in Table~\ref{tab:tags} is the result using nouns and adjectives parsed from captions using NLTK~\cite{loper2002nltk} toolkit as target concepts. This leads to an obvious improvement over the baseline (without CTN): CIDEr $120.9$ \vs $114.8$. 
We also attempt to leverage all tokens from the captions as concept targets in case of omitting essential words during parsing (see ``CAP.$^{\spadesuit}$''), which brings further incremental improvement and yield best result. Although using all tokens in the caption might inevitably introduce more noisy or irrelevant words, \eg, connection and stop words, it also broadens the semantic concepts vocabulary as some rare entities/attributes might be missed using just keywords. 

\begin{table}[t]
    \centering
    \renewcommand{\arraystretch}{1.1} 
    \scalebox{0.95}
    {\small
    \begin{tabular}{p{27.5mm} c c c c c }
    \toprule
    \\[-3.5ex]
   \multirow{2}{*}{\textbf{Methods}} & \multicolumn{4}{c}{{\texttt{\textbf{\ \ \ \ Cross-Entropy Loss}}}} \\
    & \multicolumn{1}{c}{B@4 } & M & R  & C     & \multicolumn{1}{c}{S}    \\	
    \hline
    \cellcolor{red!3}ViT/B&   \cellcolor{gray!5}$33.9$ & $27.8$ & $56.4$ &  \cellcolor{gray!5}$114.8$ & $21.3$ \\
    \cellcolor{red!3}ViT/B${\tiny  \scalebox{1.1}{+} \  \color{darkred}{\text{\textbf{KD}}}}$ &   \cellcolor{gray!20}$35.4$ & $28.5$ & $57.5$ & \cellcolor{gray!40}$120.0$ & $21.7$   \\
    \cellcolor{red!3}ViT/B${\tiny \scalebox{1.1}{+} \ \text{CTN-TAG}}$ &    \cellcolor{gray!15}$35.2$ &  $28.0$ & $57.0$ &	\cellcolor{gray!25}$117.1$ & $21.4$ \\
    \cellcolor{red!3}ViT/B${\tiny \scalebox{1.1}{+} \ \text{OD-TAG}}$ &    \cellcolor{gray!10}$34.3$ &  $28.2$ & $57.4$	& \cellcolor{gray!25}$117.4$ & $21.7$\\
    \cellcolor{red!3}ViTCAP${\tiny \scalebox{1.1}{+} \ \text{CTN-TOK}{\color{black}}}$ &    \cellcolor{gray!25}$35.7$ &  $28.8$ & $57.6$	& \cellcolor{gray!35}$121.8$ & $22.1$\\
    \cellcolor{red!3}ViTCAP${\tiny \scalebox{1.1}{+} \ \text{CTN-TOK}  \scalebox{1.1}{+} \ {\color{ddarkgreen}{\text{\textbf{PRE}}}} \scalebox{1.1}{+} {\color{darkred}{\text{\textbf{KD}}}} }$ &  \cellcolor{gray!35}$36.3$ &  $29.3$ & $58.1$ & \cellcolor{gray!45}$125.2$ & $22.6$   \\
    \bottomrule
    \end{tabular}
    }
    \vspace{2mm}
    \caption{\small Comparisons of \vitcapp with or without knowledge distillation, large-scale pre-training and with CTN. Performances are reported on COCO-caption Karpathy split optimized by cross-entropy loss. $_{+ \text{OD-TAG}}$ indicates the result using the detector produced off-the-shelf tags as~\cite{li2020oscar}. $_{+ \text{CTN-TOK}}$ is the result of \vitcapp using the initialization after first-stage concept classification. $_{\color{darkred}{\text{\textbf{KD}}}}$ and $_{\color{darkgreen}{\text{\textbf{PRE}}}}$ are results obtained with masked token classification distillation and pre-training at scale respectively.
    }
  \label{tab:ablation}
  \vspace{-4mm}
\end{table}
We then experiment with using the detectors in~\cite{zhang2021multi} and~\cite{anderson2018bottom} to produce image-level tags as target concepts. We observe that using the detector of VinVL yields better performances than BUTD, \ie, $119.7$ \vs $117.4$ CIDEr scores. This is mainly because of the more diverse collection of semantic concepts involved in~\cite{zhang2021multi} than BUTD~\cite{anderson2018bottom}. 
The second last row is the experiment where the model is firstly trained using VinVL tags on large scale dataset (in the first stage), and then using the caption tokens during the second stage of captioning. This indicates that, when no captions are attainable, it is also viable to leverage detector-produced tags to improve the performance.

\begin{table*}[t!]
	\begin{minipage}{0.28\linewidth}
    \centering
    \renewcommand{\arraystretch}{1.05} 
    \scalebox{0.98}
    { 
    \small
    \begin{tabular}{l|c}
    \toprule
    \multirow{2}{*}{\textbf{Methods}}  & \multicolumn{1}{c}{{ \texttt{\textbf{CC-3M dev}}}}  \\ 
    & {CIDEr} \\
    \hline 
       \cellcolor{red!3}FRCNN~\cite{changpinyo2019decoupled} & $89.2$ \\
       \cellcolor{red!3}Ultra~\cite{changpinyo2019decoupled} & $93.7$ \\ 
       \cellcolor{red!3}ViLT-CAP~\cite{kim2021vilt}$^{\color{black}{\spadesuit}}$ & $83.8$\\ 
       \cellcolor{red!3}VinVL~\cite{zhang2021multi}$^{\color{black}{\spadesuit}}$ & $103.4$ \\ 
       \cellcolor{red!3}CC-$3$M~\cite{changpinyo2021conceptual} & $100.9$ \\
       \cellcolor{red!3}CC-$12$M~\cite{changpinyo2021conceptual} & $105.4$ \\ 
       \hline
       \cellcolor{blue!3}ViTCAP  &  $\textbf{108.6}_\text{\color{darkgreen}\textbf{ +3.2}}$ \\
    \bottomrule
    \end{tabular}
    }
    \caption{\small
    Performances of \vitcapp model on Conceptual Captions (Google-CC 3M dev-split)~\cite{sharma2018conceptual} benchmark. We compare with the baseline methods FRCNN~\cite{changpinyo2019decoupled}, Ultra~\cite{changpinyo2019decoupled} and~\cite{changpinyo2021conceptual}. The ViLT-CAP$^{\color{black}{\spadesuit}}$ and VinVL represent our reproduced results with pre-trained checkpoint from~\cite{kim2021vilt} and~\cite{zhang2021multi}.
    }
    \label{tab:googlecc}
	\end{minipage} \hfill
	\begin{minipage}{0.68\linewidth}
    \centering
    \setlength{\tabcolsep}{4pt} 
    \renewcommand{\arraystretch}{1.05} 
    {
     \small
    \begin{tabular}{l|cc|cc|cc|cc}
    \toprule
    & \multicolumn{8}{c}{\texttt{\textbf{nocaps validation set}}} \\
    \cline{2-9} 
    \multicolumn{1}{l|}{\textbf{Methods}} & \multicolumn{2}{c|}{\textbf{in-domain}} & \multicolumn{2}{c|}{\textbf{near-domain}} &  \multicolumn{2}{c|}{\textbf{out-of-domain}} & \multicolumn{2}{c}{\textbf{overall}}\\ 
     & C & S & C &  S  & C & S & C & S \\ 
    \hline 
    \transparent{0.4}Human & \transparent{0.4}84.4 & \transparent{0.4}\textbf{14.3} & \transparent{0.4}85.0 & \transparent{0.4}\textbf{14.3} & \transparent{0.4}\textbf{95.7} & \transparent{0.4}\textbf{14.0} & \transparent{0.4}87.1 & \transparent{0.4}\textbf{14.2} \\
    \hline
    \cellcolor{red!3}UpDown~\cite{agrawal2019nocaps} & $78.1$ & $11.6$ & $57.7$ & $10.3$ & $31.3$ & $8.3$ & $55.3$ & $10.1$ \\
    \cellcolor{red!3}UpDown + CBS  & $80.0$ & $12.0$ & $73.6$ & $11.3$ & $66.4$ & $9.7$ & $73.1$ & $11.1$ \\
    \cellcolor{red!3}UpDown + ELMO + CBS  & $80.0$ & $12.0$ & $73.6$ & $11.3$ & $66.4$ & $9.7$ & $73.1$ & $11.1$ \\
    \cellcolor{red!3}OSCAR~\cite{li2020oscar} & $79.6$ & $12.3$ & $66.1$ & $11.5$ & $45.3$ & $9.7$ & $63.8$ & $11.2$ \\
    \cellcolor{red!3}OSCAR + CBS & $83.4$ & $12.0$ & $81.6$ & $12.0$ & $77.6$ & $10.6$ & $81.1$ & $11.7$ \\
    \cellcolor{red!3}VIVO ~\cite{hu2020vivo} & $90.4$ & $13.0$ & $84.9$ & $12.5$ & $83.0$ & $10.7$ & $85.3$ & $12.2$ \\
    \cellcolor{red!3}VIVO + CBS & $92.2$ & $12.9$ & $87.8$ & $12.6$ & $87.5$ & $11.5$ & $88.3$ & $12.4$ \\
    \hline
    \cellcolor{blue!3}\vitcap  & $\textbf{99.3}$ & $13.2$ & $90.4$ & $12.9$ & $78.1$ & $11.9$ & $89.2$ & $12.7$ \\
    \cellcolor{blue!3}\vitcap + CBS & $98.7$ & $13.3$ & $\textbf{92.3}$ & $13.3$ & ${95.4}$ & ${12.7}$ & $\textbf{93.8}$ & $13.0$ \\ [-6pt]
     $_{\color{black}\Delta}$ & $_\text{\color{darkgreen}\textbf{+6.5}}$ & $_\text{\color{darkgreen}\textbf{+0.4}}$ & $_\text{\color{darkgreen}\textbf{+4.5}}$ & $_\text{\color{darkgreen}\textbf{+0.7}}$ & $_\text{\color{darkgreen}\textbf{+7.9}}$ & $_\text{\color{darkgreen}\textbf{+1.2}}$ &
     $_\text{\color{darkgreen}\textbf{+5.5}}$ &
     $_\text{\color{darkgreen}\textbf{+0.6}}$ \\
    \bottomrule
    \end{tabular}
    }
    \vspace{1mm}
    \caption {\small Performances of \vitcapp in nocaps validation split. We compare our \vitcapp with previous state-of-the-art models at \textbf{``in-domain''}, \textbf{``near-domain''} and \textbf{``out-of-domain''}. Results are reported with constrained beam search (CBS) decoding~\cite{anderson2016guided}.}
    \label{tab:arch}
	\end{minipage} \ \ \
	\vspace{-4mm}
\end{table*}

\vspace{1mm}
\noindent \textbf{Effect of Different Modules.} \label{sec:module} In Table~\ref{tab:ablation}, we show in details the independent performance gains from each design, \textit{viz.}, with or without concept tokens, masked token distillation loss, pre-training and the combinations of them. 
We report the result of the baseline model 
which reaches CIDEr scores $114.8$ on COCO caption dataset. 
With the aim of isolating the performance gain from concept tokens, we first decode the image-level semantic concepts and store them as offline tags for the captioning task. We then follow~\cite{li2020oscar} to tokenize them and concatenate the tag embedding with visual features for captioning task. This allows us to directly compare the effect of CTN-produced concepts with detector tags without the concept classification initialization.
Adopting the explicit tags predicted by the CTN leads to obvious improvements: $2.3$ higher CIDEr and $1.3$ higher BLEU@$4$ scores, reaching similar results with that using VinVL's detector tags directly (see ViT/B$_\text{+OD-TAG}$): $117.4$ \vs $117.1$ CIDEr scores. This proves that our generated semantic concepts play a significant role in the captioning task and have a similar effect as the VinVL's detector tags. Next, we apply the pre-trained weights after the concept classification to initialize the \vitcap for the captioning task, and find further improvement (see ViTCAP${_\text{+CTN-TOK}}$). 
This proves that both the predicted concept tokens and the concept classification training are beneficial for captioning tasks. 
For the knowledge distillation experiment, we use the VinVL-base~\cite{zhang2021multi} optimized on COCO-caption dataset as the Teacher and keep it frozen during distillation.
The application of KD on masked token prediction (ViT/B${_\text{+\color{darkred}{\textbf{KD}}}}$) is also evidently helpful: there is an over $5.0$ CIDEr scores improvement over the baseline. Note that the KD objective is only applied in the downstream for the \vitcap baseline after VLP for fair comparison with previous works. 
Finally, by pre-training the \vitcap with large scale VL corpus continuously contributes to the results.

\noindent \textbf{Performances on other Benchmarks.}
\label{sec:otherdata} To evaluate the generalizability of ViTCAP, we continue to expand the testbeds to other challenging captioning benchmarks, \ie, Google-CC~\cite{sharma2018conceptual} and nocaps~\cite{agrawal2019nocaps} datasets. For the Google-CC dataset, we train the \vitcap on the training split, which consists of ${\sim}3.3$M image-caption pairs, and test it on the dev split. We follow the same training protocols as previously mentioned and optimize the \vitcap for $120$ epochs.  Following previous works, we evaluate the performances using the CIDEr metric and Table~\ref{tab:googlecc} shows the results of \vitcap compared with previous captioning models. In particular, \vitcap achieves the state-of-the-art results CIDEr $108.6$ scores (without the knowledge distillation), surpassing all detector-based captioning models. CC-$12$M is the model trained with $12$M image-caption pairs~\cite{changpinyo2021conceptual}. Again, when evaluating on nocaps dataset, \vitcap  shows promissing results across all in-domain, near-domain, and out-of-domain splits. For example, \vitcap achieves $98.7$ and $93.8$ CIDEr scores on in/out-domain splits, $6.5$ and $5.5$ higher than the VIVO~\cite{hu2020vivo}, which exploits OpenImage~\cite{kuznetsova2020open} dataset to learn semantic concepts for captioning task. 
The great generalization ability of \vitcap can be partly ascribed to its ability to recognize expansive semantic concepts extracted from the open-form captions. Compared to predicting the pre-defined tags as in the detector, the usage of caption extracted concepts largely expands the concept vocabulary. This provides the \vitcap with robust and broad concept tokens, which is essential for the images with novel concepts.

\begin{figure}[t!]
  \begin{center}
  \includegraphics[width=.46\textwidth]{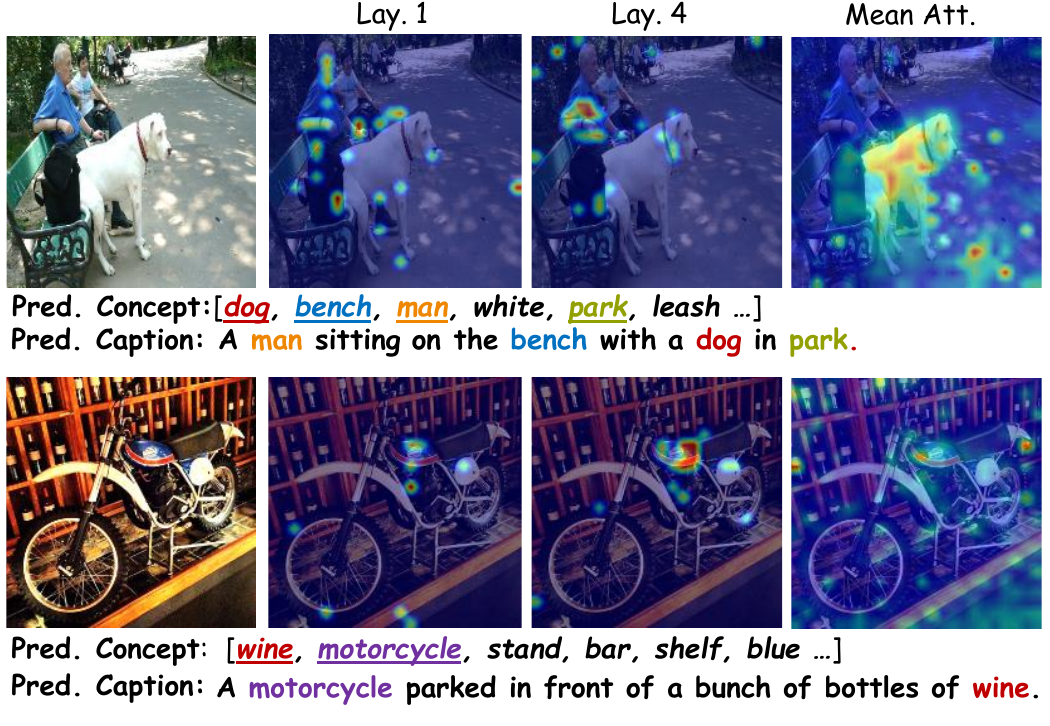}
  \end{center}
  \vspace{-6mm}
    \caption{\small Visualization of the attention maps from \vitcapp and its produced concepts\&captions. ``\rule{6mm}{0.15mm}'' refers to the concepts appear in captions. Best viewed in color.
    }
  \label{fig:qualitative_main}
  \vspace{-2mm}
\end{figure}

\vspace{1mm}
\noindent \textbf{Qualitative Examples.} We show visualization examples of the attention maps from \vitcap in Figure~\ref{fig:qualitative_main} together with their generated concepts\&captions. Interestingly, we observe obvious correlations between the attended regions across different layers and predicted concepts. For example, ``{\color{red} \texttt{\textbf{dog}}}'' is notably highlighted according to the mean-averaged attention maps, yet the ``{\color{orange} \texttt{\textbf{man}}}'' is more attended in shallower transformer blocks. We conjecture that instead of relying on an object detector to glean object locations, training the detector-free VL model properly via image-text supervisions might potentially lead to a strong grounding model.

\section{Conclusion}
In this paper, we propose the ViTCAP, a detector-free image captioning model in the full transformer architecture fashion. Compared with existing captioning models, \vitcap can be trained in an end-to-end fashion without intermediate regional operations using grid representations. Our proposed Concept Token Network learns broad semantic concepts and encodes them as the concept tokens that largely benefit the captioning task on a series of challenging captioning benchmarks. Extensive experiments indicate that \vitcap achieves competing performances, approaching most detector-based models. We anticipate that ViTCAP will lead to more future works in building efficient Vision and Language models.


\vspace{1mm}
\noindent \textbf{Acknowledgement.}
This work was supported by the National Science Foundation under Grant CMMI-1925403, IIS-2132724 and IIS-1750082. 


\def\abstract
 {%
 \iftoggle{cvprpagenumbers}{}{%
\thispagestyle{empty}
 }
 \centerline{\large\bf Supplementary Materials}%
\vspace*{12pt}%
 \it%
 }

\def\endabstract
 {
 \vspace*{12pt}
 }

\vspace{8mm}
\begin{abstract}
In this supplementary materials, we provide additional details about experimental settings, and then further compare effect of different semantic concept sources, more ablative studies regards training, different architectural instantiations, and further showcase more qualitative examples of predicted semantic concepts. 

\end{abstract}

\begin{table}[h!]
\begin{center}
\begin{tabular}{c@{\hspace{3pt}}|c@{\hspace{3pt}}|c|c|c@{\hspace{3pt}}}
\toprule
Source & VG~\cite{krishna2017visual} & COCO~\cite{lin2014microsoft} & CC ~\cite{changpinyo2021conceptual}& SBU~\cite{ordonez2011im2text} \\
\midrule
Image & 108K & 113K & 3.1M & 875K \\
Text & 5.4M & 567K & 3.1M & 875K \\
\bottomrule
\end{tabular}
\end{center}
\caption{Statistics of the VL pre-training datasets. }
\label{tab:VLcorpus}
\vspace{-3mm}
\end{table}

\section{Pre-training VL Corpus}
As previous works in~\cite{zhang2021multi}, we carry out the pre-training of ViTCAP on the aggregation of several common datasets, which include COCO~\cite{lin2014microsoft}, Conceptual Caption~\cite{changpinyo2021conceptual}, SBU Captions~\cite{ordonez2011im2text}, and Visual Genome~\cite{krishna2017visual}. We have the detailed  statistics of the aggregated datasets in Table~\ref{tab:VLcorpus}. In total, we use $4.2$ millions of images and $9.9$M captions for the pre-training. Following~\cite{lu202012}, we de-duplicate images that exist in both pre-training corpus and COCO Karpathy testing splits for fair comparisons.

\section{Ablative Studies}
This section further presents additional ablative studies about ViTCAP, which includes: some examples and basic statistics about semantic concepts, the effect of different concept sources, 
results of different concept classification losses, different other training strategies.

\begin{figure}[h]
 \centering
 \includegraphics[width=.46\textwidth]{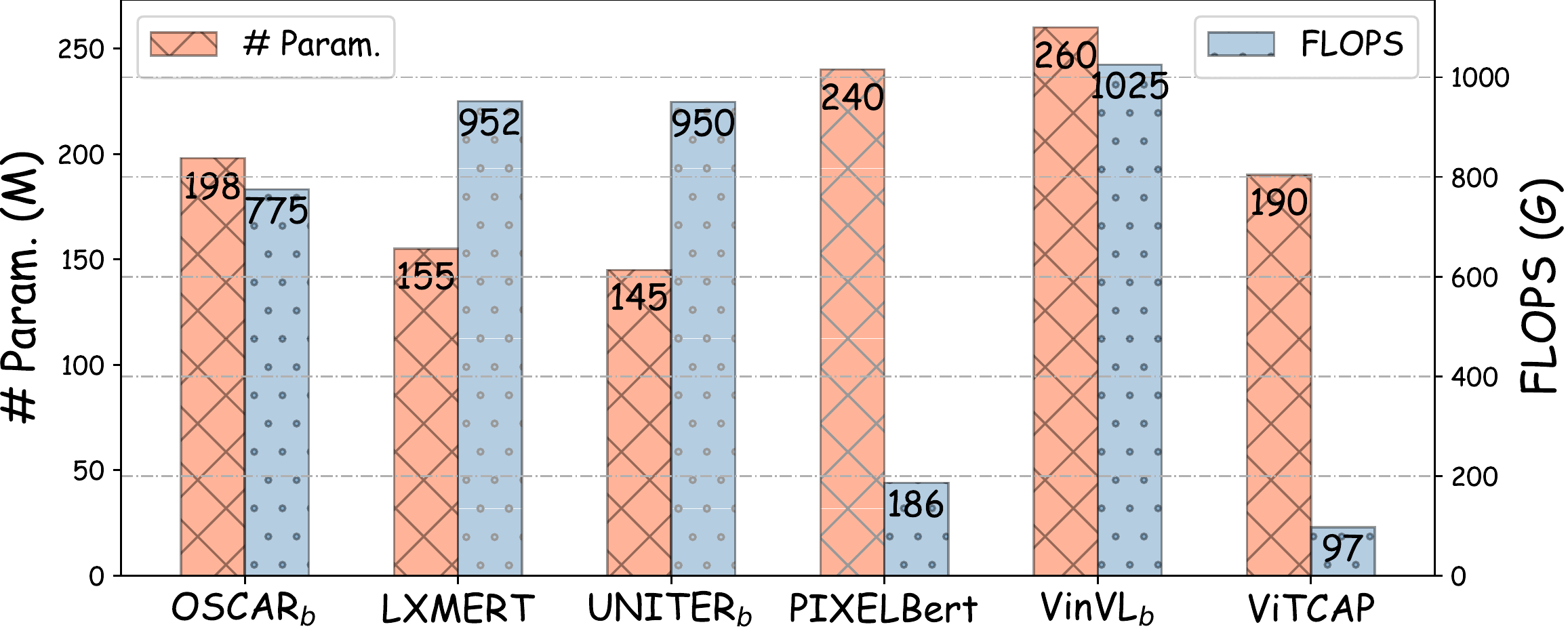}
 \caption {\small Inference speed in FLOPs (in G), number of parameters (in M) of multiple VL models and \vitcapp\!\!.}
 \vspace{-1mm}
\label{fig:flops}
\end{figure}

\vspace{1mm}
\noindent\textbf{Examples and Stats of Concepts.}
In practice, we experiment with utilizing semantic concepts gleaned from 1). open-form image captions by language parsing (or simple as using all tokens as classification ground-truth) or 2).  an object detector. 

\begin{figure}[t]
\centering
 \includegraphics[width=1\linewidth]{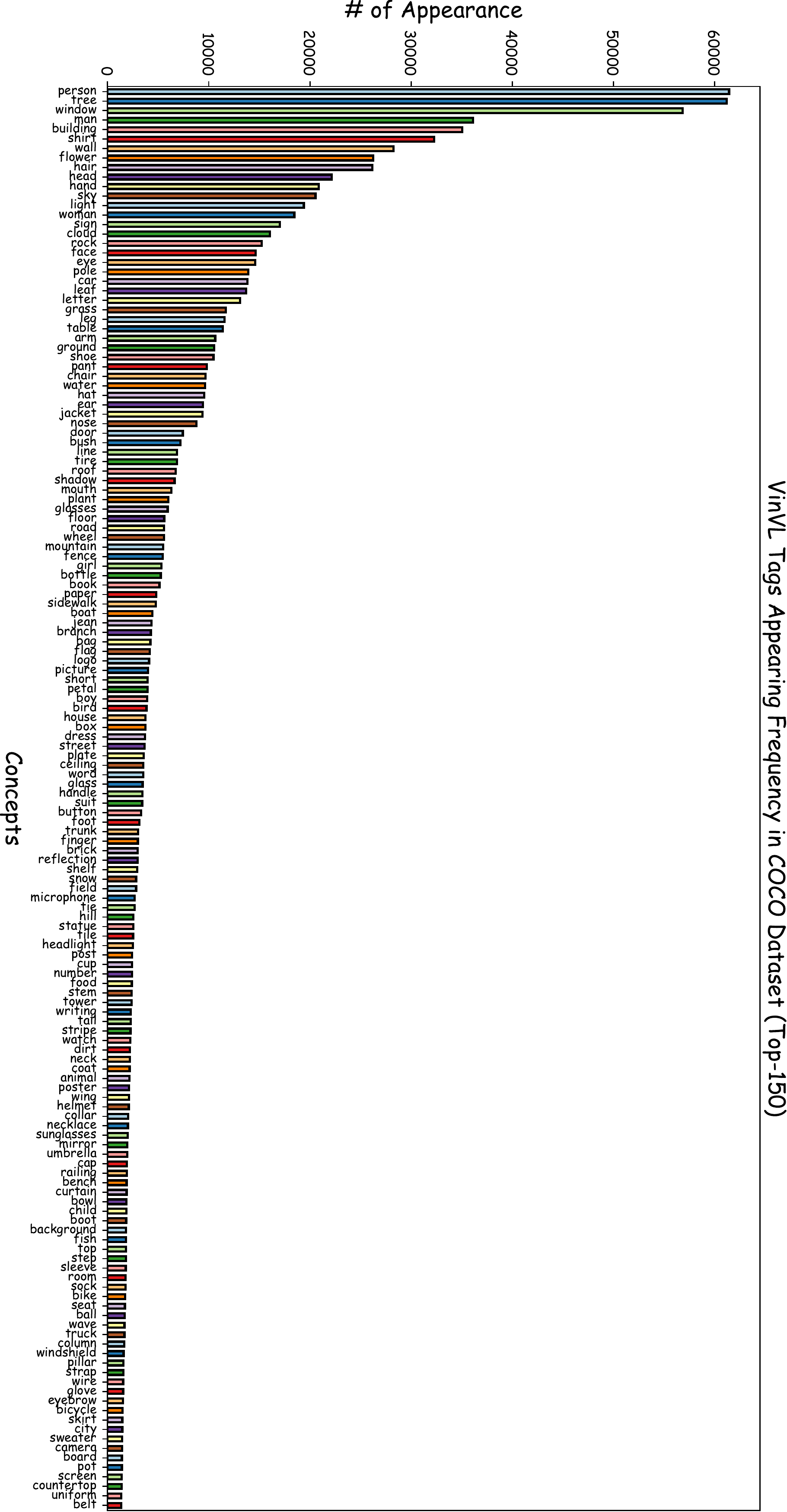}
\vspace{-2mm}
\caption{Top-$150$ most frequently appeared semantic concepts produced by VinVL's object detector. The produced tags are severely long-tail distributed and certain concepts dominates across all samples. This arises the necessity to apply focal loss as countermeasure. }
\label{fig:stats}
\end{figure}

As previously mentioned, we notice that the concepts from both sides are all severely long-tailed distributed (an example of the detector-produced concept distribution is shown in Figure~\ref{fig:stats}). Notably, certain concepts appear more frequently across the whole COCO training split, \eg, ``\texttt{person}'', ``\texttt{tree}'', ``\texttt{window}'' obviously exist far more frequent than the remaining.  We also resort to different object detectors to acquire high-quality semantic concepts, \ie, a ResNet$_{101}$ base Faster-RCNN~\cite{anderson2018bottom} that has been pre-trained on Visual-Genome dataset~\cite{krishna2017visual} (denoted as BUTD), and a ResNext$_{152}$ based modified Faster-RCNN detector with broader categories of the visual attribute as detection targets (denoted as VinVL). These detector-produced image-level tags are actually accurate with less noise than in captions, but they also require a pre-defined categorical dictionary with a fixed set of concepts. This largely limits the scope of their applications.

In Figure~\ref{fig:flops}, we present the inference speed and the number of learnable parameters of prevailing detector-based VL models compared with \vitcap\!\! Notably, with on-par parameters, \vitcap consumes only $\sim10\%$ FLOPs of the prevailing VL models ($97$G for \vitcap \vs $1,025$G for VinVL).

\vspace{1mm}
\noindent\textbf{More About Concept Sources.}
Open-form captions are the most ideal source to obtain semantic concepts as they naturally carry abundant semantic concepts with no vocabulary limitation. Notwithstanding that most of these descriptions can be noisy, inaccurate, and incomplete. In practice, we leverage different ways to extract the concepts from them by 1) using the NLTK~\cite{loper2002nltk} toolkit and parsing out only the nouns and adjectives as the semantic concepts for the classification task (see ``CAPTION'' baseline in main paper); 2) we also simply attempt to leverage all tokens from the captions as concept targets in case of omitting essential words during parsing (see ``$^{\spadesuit}$'' in main paper).
We first extract these tags as ``\textit{off-the-shelf}'' annotations for the concept classification task and then apply the initialization of ViTCAP after the first stage of training for the joint captioning training. Note that we conduct and compare all these ablations without VL pre-training. It is beneficial to further adopt the concept classification loss during the joint training, as the semantic concepts in the COCO-caption dataset vary with the concept classification dataset. Also, captions in these two domains might vary from the aspect of textual styles: for example, length of captions, the use of synonyms, cognate and conjugate words, or various tenses.

\begin{table}[h]
\centering
\setlength{\tabcolsep}{4pt} 
\renewcommand{\arraystretch}{1.15} 
\scalebox{0.99}{
\small
\begin{tabular}{l|c|ccccc}
\toprule
& \multicolumn{5}{c}{{ \texttt{\textbf{COCO Captioning}}}} \\ 
& EPOCH & {B@4} & {M} & {R} & {C} & {S} \\
\hline 
{\transparent{0.4} Baseline} & {\transparent{0.4} -} & {\transparent{0.4} $33.9$} &
{\transparent{0.4} $27.8$} & {\transparent{0.4} $56.4$} & {\transparent{0.4} $114.8$} & {\transparent{0.4} $21.3$} \\
{\transparent{0.4} VinVL-Tag} & {\transparent{0.4} -} & {\transparent{0.4} $35.4$} & {\transparent{0.4} $28.1$} & {\transparent{0.4} $57.2$} & {\transparent{0.4} $117.7$} &  {\transparent{0.4} $21.3$} \\
\hline
BCE$_\text{Tag}$  & $10$ & $33.9$ & $27.9$ & $56.5$ & $115.0$ & $21.4$\\ 
FOCAL$_\text{Tag}$ & $10$ & $35.2$ & $28.0$ & $57.0$ & $117.1$ & $21.4$\\ 
FOCAL$_\text{Tag+Init}$ & $10$ & $36.0$ & $28.4$ & $57.5$ & $120.5$ & $22.0$\\ 
FOCAL$_\text{Init}$ & $10$ & $35.0$ & $28.2$ & $57.1$ & $118.0$ & $21.6$\\ 
FOCAL$_\text{Tag+Init}$ & $40$ & $35.9$ & $28.4$ & $57.6$ & $121.1$ & $22.1$ \\ 
\bottomrule
\end{tabular}
}
\vspace{2mm}
\caption{\small Performances of ViTCAP using focal loss, binary classification loss as concept classification training target. 
}
\label{tab:losses}
\vspace{-4mm}
\end{table} 

\vspace{1mm}
\noindent\textbf{Concept Classification Training.}
We now study the effect of different losses for the concept classification task, namely binary cross-entropy loss and focal loss, and the effect of the initialization after the classification training. 
The extremely imbalanced sample distribution usually leads to sub-optimal classification performances, as also studied in previous works like face recognition~\cite{zhang2017range,ma2020learning} and object detection~\cite{li2020overcoming, ouyang2016factors}, etc. As countermeasures, there exist works designing advanced losses~\cite{lin2017focal,zhang2017range} re-weighting different samples. In Table~\ref{tab:losses}, we list the performances of ViTCAP using different losses. In specific, the top-two rows are the baseline results 1). Baseline: vanilla Encoder-Decoder architecture without CTN branch, and 2). Encoder-Decoder architecture using VinVL's OD tags as~\cite{li2020oscar}. ``$_\text{Tag}$'' denotes the results are reported using concepts as the offline tags without concept classification \& its initialization. We observe that by applying the BCE loss trained offline concepts as offline tags, the results are only incrementally improved over the baseline, and it still shows a great performance gap \textit{w.r.t.} the VinVL's tag. Notably, using focal loss obviously improves the quality of produced concepts, reaching $117.1$ CIDEr scores. To this end, we apply the concept classification pre-trained initialization, and this
further improves the performances to a great extent. It is discernible that the experiment ``$_\text{Init}$'' gives worse result than the ``$_\text{Tag+Init}$''. This validates that both the concept classification task and the predicted concepts are helpful for the captioning task. Results show that they are complementary to each other.

\begin{table}[h]
\centering
\setlength{\tabcolsep}{4pt} 
\renewcommand{\arraystretch}{1.15} 
\scalebox{0.99}{
\small
\begin{tabular}{l|ccccc}
\toprule
\multirow{2}{*}{Tokenization} & \multicolumn{5}{c}{{ \texttt{\textbf{COCO Captioning}}}} \\ & {B@4} & {M} & {R} & {C} & {S} \\
\hline 
Caption Tokenizer & $35.5$ & $28.5$ & $57.5$ & $119.7$ & $21.8$ \\ 
Classifier Tokenizer & $35.6$ & $28.4$ & $57.4$ & $119.8$ & $21.8$ \\ 
Independent Tokenizer & $35.9$ & $28.5$ & $57.6$ & $120.1$ & $21.9$ \\ 
\bottomrule
\end{tabular}
}
\vspace{2mm}
\caption{\small Performances of ViTAP using different strategies for concept tokenization. 
}
\label{tab:tokenizer}
\vspace{-4mm}
\end{table} 

\noindent\textbf{Representing Concepts as Tokens.} There are multiple ways to encode the predicted concepts as continuous embedding for the decoding stage. We study three different ways of encoding and present the results in Table~\ref{tab:tokenizer}, namely, 1). use the tokenizer for captioning, 2). use the concept classifier's tokenizer (in concept classification, we simply use the BERT tokenizer to encode the semantic concepts), 3). use an independent and untrained tokenizer. Though in practice, all three tokenizers are implemented based on the BERT tokenizer~\cite{devlin2018bert}, the embeddings from the three are entirely different. From the results, we observe a fairly negligible performance gap: using an independent tokenizer only yields a $0.4$ higher CIDEr score. Though adopting an independent tokenizer yield the best result, it introduces additional parameters and thus we choose to share the tokenizer for captioning instead.


\begin{table}[h]
\centering
\setlength{\tabcolsep}{4pt} 
\renewcommand{\arraystretch}{1.15} 
\scalebox{0.99}{
\small
\begin{tabular}{l|ccccc}
\toprule
\multirow{2}{*}{\textbf{}} & \multicolumn{5}{c}{{ \texttt{\textbf{COCO Captioning}}}}  \\ 
& {B@4} & {M} & {R} & {C} & {S} \\
\hline 
GT Concepts & $35.5$ & $28.4$ & $57.3$ & $119.1$ & $21.7$ \\ 
GT + PRED. Concepts & $35.2$ & $28.5$ & $57.3$ & $119.2$ & $21.8$\\ 
PRED. Concepts & $36.1$ & $28.6$ & $57.6$ & $120.6$ & $21.7$\\ 
\bottomrule
\end{tabular}
}
\vspace{2mm}
\caption{\small Performances of VitCAP using either ground-truth concepts for captioning, the concept network predicted concept tokens or the mixture of them during training. 
}
\label{tab:concept}
\end{table}

We experiment with different ways to train with the concept tokens. In Table~\ref{tab:concept}, we list the results of training using GT semantic concepts encoded as tokens, GT concepts mixed with predicted concepts, and fully predicted concepts.

\begin{figure}[h]
\begin{center}
\includegraphics[width=.46\textwidth]{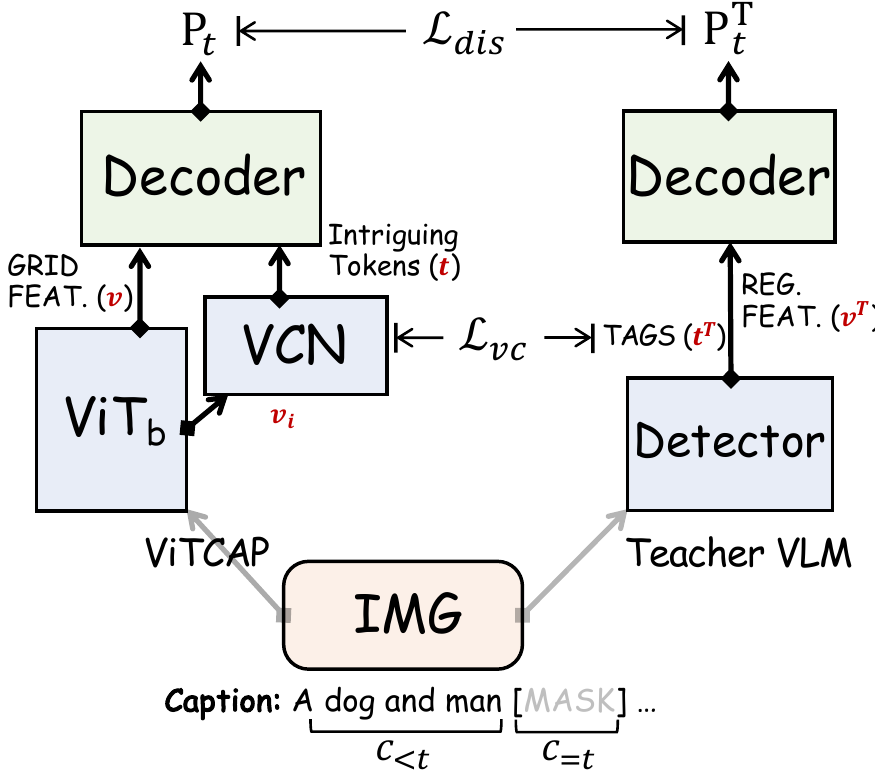}
\end{center}
\vspace{-6mm}
\caption{\small The overall training paradigm of ViTCAP can be understood as the knowledge distillation procedure where a detector-based Teacher VLM to assist the training of ViTCAP as a knowledge distillation paradigm. The CTN branch in ViTCAP learns to predict the semantic concepts as conceptual tokens for captioning. }
\vspace{-2mm}
\label{fig:distillation}
\end{figure}

We find that by using the predicted concepts for training leads to optimal results. This is mostly because the pre-trained CTN can already produce reasonable concepts at the captioning fine-tuning stage.

\begin{table}[h]
\centering
\setlength{\tabcolsep}{6pt} 
\renewcommand{\arraystretch}{1.15} 
\scalebox{0.99}{
 \small
\begin{tabular}{l|ccccc}
\toprule
\multirow{2}{*}{\textbf{Architecture}} & \multicolumn{5}{c}{{ \texttt{\textbf{COCO Captioning}}}}  \\ 
& {B@4} & {M} & {R} & {C} & {S} \\
\hline 
SIN-TOW$_{32\times32}$ & $32.5$ & $27.1$ & $55.4$ & $109.5$ & $20.2$ \\ 
$_{+\text{EFF. OD-Tags}}$ & $32.8$ &  $27.4$ & 	$55.5$ & $110.9$ &	$20.6$ \\
$_{+\text{VinVL-Tags}}$ &  $33.5$ & $27.8$ & $56.1$ & $114.6$ & $21.1$\\
\hline
ENC-DEC$_{32\times32}$ & $33.4$ & $27.5$ & $56.0$ & $112.1$ & $20.6$ \\
$_{+\text{EFF. OD-Tags}}$ & $33.8$ & $27.9$ & $56.4$ & $114.6$ & $21.3$  \\
$_{+\text{VinVL-Tags}}$ & $34.4$ & 	$27.9$ & $56.6$ & $115.8$& $21.1$\\
$_{+\text{ViTCAP-Tags}}$ & $34.0$ &	$27.7$ & $56.3$ & $114.2$ & $20.8$\\
\hline
SIN-TOW$_{16\times16}$ & $33.8$ & $27.8$ &	$56.2$ & $113.9$ & $21.0$\\ 
$_{+\text{EFF. OD-Tags}}$ & $33.8$ & $27.9$ &	$56.4$ &	$114.6$ &	$21.3$ \\
$_{+\text{VinVL-Tags}}$ & $34.3$ &	$28.2$ & $56.7$ & $117.4$	& $21.7$ \\
\hline
ENC-DEC$_{16\times16}$ & $33.9$ &$27.8$ &	$56.4$ &	$114.8$ &	$21.3$ \\
$_{+\text{VinVL-Tags}}$ & $35.4$ &	$28.1$ & $57.2$ & 	$117.7$ &	$21.3$\\
$_{+\text{ViTCAP-Tags}}$ & $35.2$ & 	$28.0$ & 	$57.0$	 & $117.1$ & 	$21.4$\\
\hline
ViTCAP & $35.7$ &  $28.8$ &  $57.6$ & $121.8$ & $22.1$\\
\bottomrule
\end{tabular}
}
\vspace{2mm}
\caption{\small
We compare different instantiations of ViTCAP with architectural variations of ViT based captioning model: single-tower (SIN-TOW), encoder-decoder structure (ENC-DEC), two-tower ViTCAP, and ViTCAP with various numbers of sharing blocks in stem image encoder. All experiments are conducted without VL pre-training and are trained by cross-entropy loss.
}
\label{tab:arch}
\end{table} 

\vspace{1mm}
\noindent \textbf{ViTCAP Architecture.} To give a more detailed explanation of the architecture of ViTCAP: it consists of a stem image encoder with $8$ transformer blocks (shared for both grid feature extractor and CTN), a CTN branch with $4$ transformer blocks, and a grid feature extractor with $4$ transformer blocks, the multi-modal module is also a $4$ transformer blocks module. When $M_1=12$, the model can be understood as consisting of two parallel branches, with one for concept prediction and one for grid representation. We does find that minimizing the shared blocks can bring extra performance gains but this inevitably increases the model size very obviously. We only adopt this two-tower design in the experiment with large scale pre-training where we follow a two-step training schema as OSCAR~\cite{li2020oscar}: we first leverage the CTN to predict the semantic concepts of all pre-training images; Then, we use these concepts as the off-the-shelf tags (similar as the object detector tags) for the pre-training.

\vspace{1mm}
\noindent \textbf{Architectural Variations.} We then experiment
with different architectural variations of ViTCAP and report their performances on COCO-caption in Table~\ref{tab:arch}. 
The baseline models include single-tower (SIN-TOW) that shares the ViT backbone for both modalities; Encoder-decoder (ENC-DEC) that use a ViT as visual encoder and 4 separate transformer blocks as modal fusion. This is similar to~\cite{kim2021vilt}, however, we modify it by using seq-to-seq attention maps for the captioning training which prevents the model from seeing bidirectional context; Two-tower (TWO-TOW) uses an independent ViT/b architecture as a conceptual token network and another architecture as the visual encoder.

\begin{table}[h!]
\begin{center}
\begin{tabular}{lc}
\toprule
Methods & SMURF \\
\midrule
w/ only periods removed  \\
VinVL & $0.66$ \\ 
M$^2$ Transformer & $0.49$ \\ 
X-Transformer & $0.51$ \\ 
ViTCAP & $0.55$ \\ 
\midrule
w/ all punctuation removed \\
VinVL & $0.59$ \\ 
M$^2$ Transformer & $0.42$\\ 
X-Transformer & $0.46$ \\
ViTCAP & $0.49$ \\
\bottomrule
\end{tabular}
\end{center}
\caption{Performance of ViTCAP comparing with previous models under SMURF~\cite{feinglass2021smurf} metric. Note that this results is evaluated using ViTCAP without pre-training.}
\label{tab:smurf}
\end{table}

\noindent\textbf{More Evaluations.} In addition to previous benchmarks, we also use the recently proposed rule-based SMURF metric which demonstrates SOTA correlation with human judgment and improved explainability. SMURF is the first caption evaluation algorithm to incorporate diction quality into its evaluation. We observe that our method preserves both semantic performance and the descriptiveness of terms used in the sentence.


\section{Discussions}

\begin{figure*}[h]
\begin{center}
\includegraphics[width=0.9\textwidth]{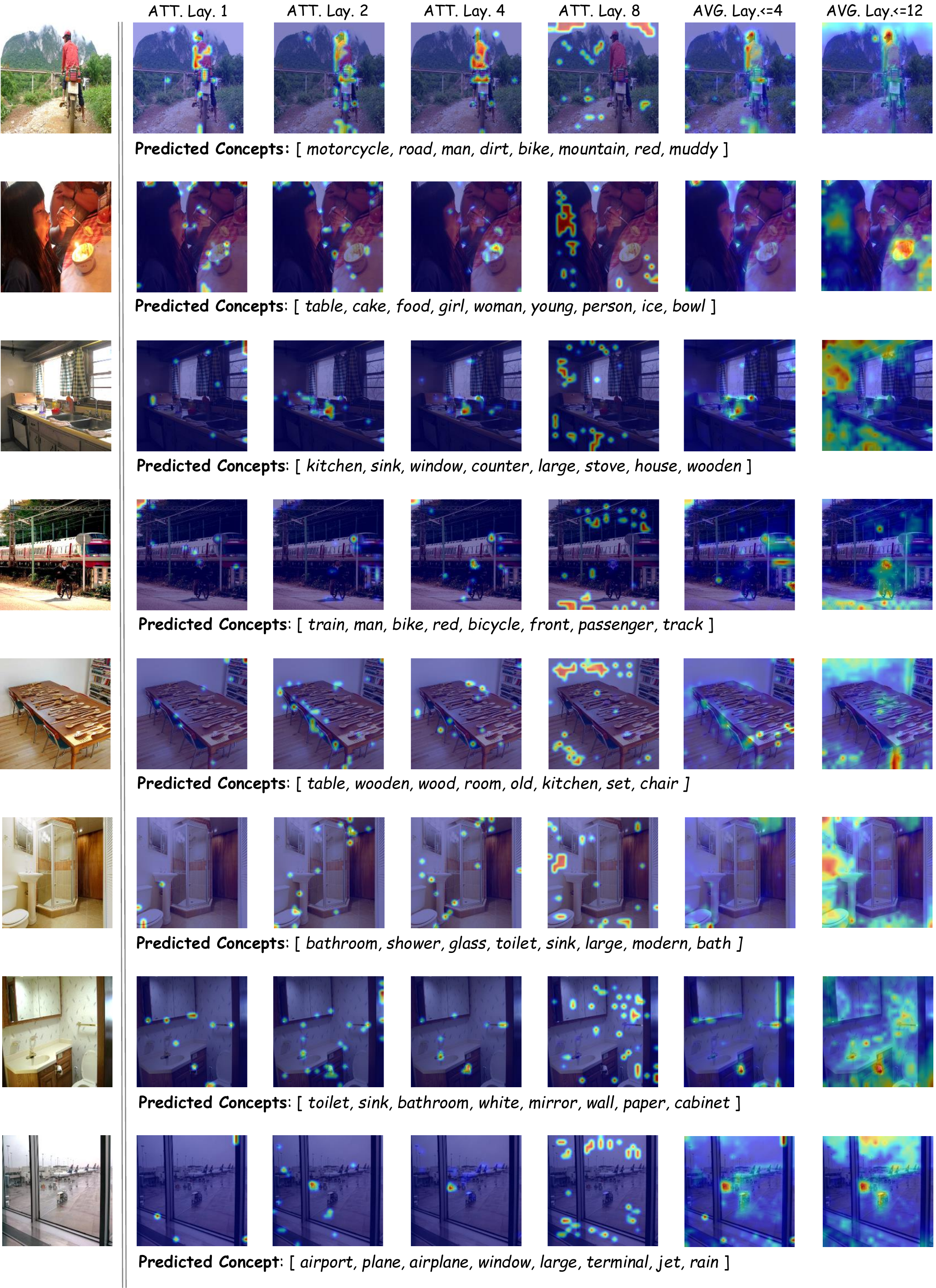}
\end{center}
\caption{\small ViTCAP produced class-agnostic attention maps and their associated semantic concepts of random images from COCO caption dataset. We exhibit attention maps of $1$, $2$, $4$, $8$th transformer blocks of ViTCAP and the mean-average attention maps of first $4$ and the entire $12$ transformer blocks (last two columns).
 }
\label{fig:qualitative}
\end{figure*}

\noindent\textbf{Qualitative Examples.}
We demonstrate more qualitative examples of the attention maps produced by ViTCAP together with their predicted semantic concepts in Figure~\ref{fig:qualitative}. 

\begin{figure*}
\begin{center}
\includegraphics[width=.75\textwidth]{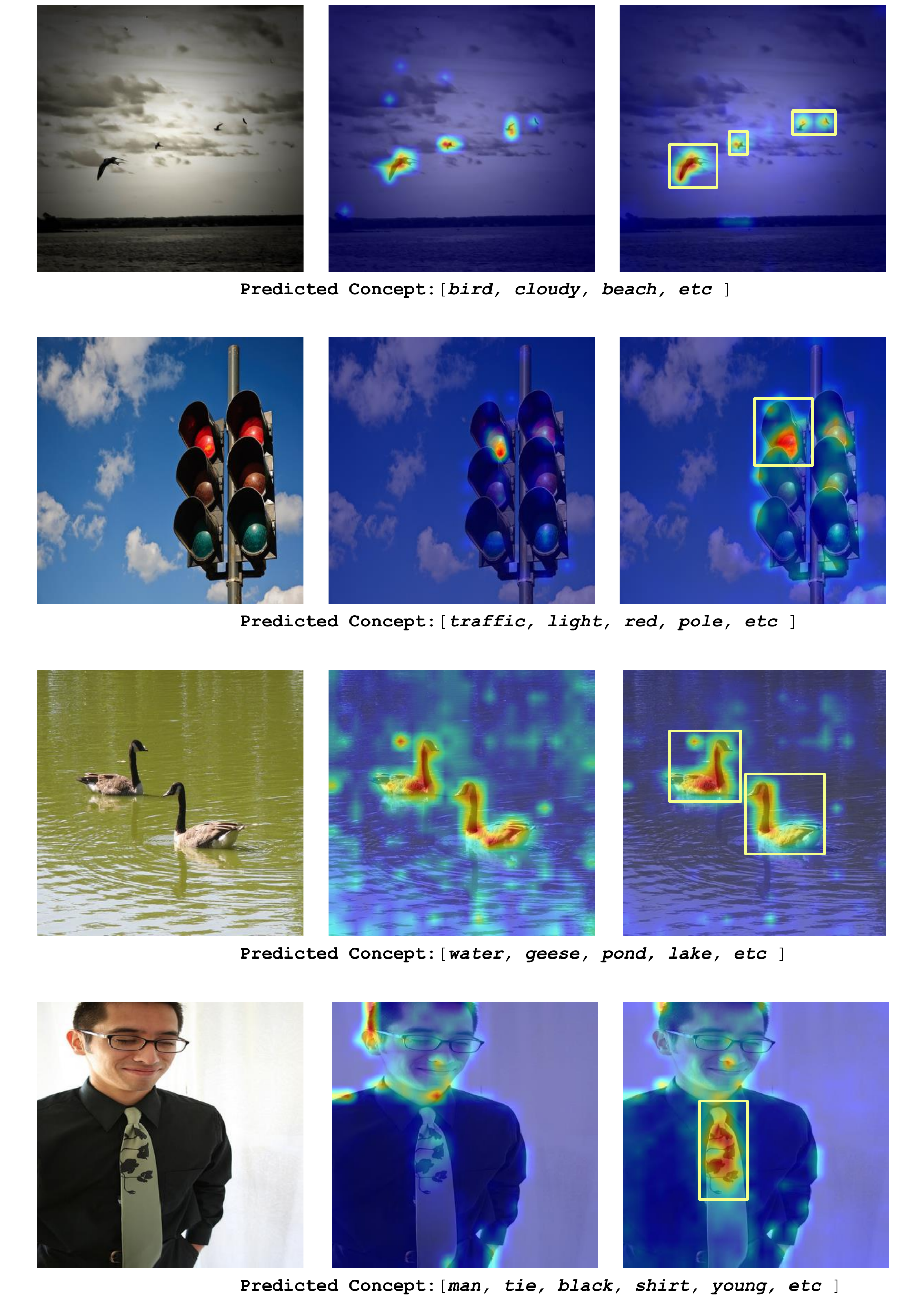}
\end{center}
\vspace{-6mm}
\caption{\small From left to right, we show the original image, average attention maps of the front 4 and 8 transformer blocks. }
\label{fig:grounding}
\end{figure*}

\vspace{1mm}
\noindent\textbf{Can ViTCAP Ground Concepts?} Interestingly, we observe that the attention maps produced from transformer blocks closely relate to the concepts and various layers have different focuses while the averaged attention maps cover broad holistic regions. We present more visualizations in Figure~\ref{fig:grounding} which contain a single object per image for more direct analysis. The topmost row is a picture with multiple ``\texttt{wild gooses}'' and all regions of them are highlighted according to the attention maps. Despite so, it seems ViTCAP suffers from identifying the clear borders of the object that it may only recognize part of the objects, \eg, ViTCAP only highlights the part of the ``\texttt{traffic light}'' and the ``\texttt{tie}''. This indicates the potential application of ViCAP for weakly supervised textual grounding tasks for the image~\cite{rohrbach2016grounding,fang2018weakly,fang2019modularized,wang2021improving} and video~\cite{huang2018finding,mithun2019weakly}.

\vspace{1mm}
\noindent\textbf{VL Distillation Schema.} Our distillation schema can be indeed viewed as an extension of the VL distillation schema, where the Student model not only mimics the predicted masked token probability but also learns from the Teacher OD's object tags. As is shown in Figure~\ref{fig:distillation}. Note that our distillation technique is only applied on the \vitcap with VL pre-training, as the teacher VL model contains knowledge acquired from large-scale pre-training and so it is unfair to compare the \vitcap with other methods without VL pre-training.

\vspace{1mm}
\noindent\textbf{Detector Tags \vs Caption Extracted Concepts.} Empirical studies show that the caption extracted concepts lead to better \vitcap\!. We conjecture that this is mainly because the captions contain much broader image concepts contained in open-form texts, yet the detector tags are pre-defined with much more limited vocabulary. However, perfectly aligned image-text pairs are not always attainable considering that most existing image-level annotations are collected from the Web. These image captions can be as noisy as alt text or short phrases, from which the extracted concepts only cover part of the image content. Thus in practice, it is also an important aspect to explore the feasibility of adopting the non-caption-extracted concepts, \eg, from an object detector as a substitution. This provides a flexible source of the concepts. 


\clearpage
{\small
\bibliographystyle{ieee_fullname}
\bibliography{egbib}
}

\end{document}